\documentclass[sn-nature]{sn-jnl}

\usepackage{natbib}
\setcitestyle{super,sort&compress}
\usepackage{graphicx}%
\usepackage{multirow}%
\usepackage{amsmath,amssymb,amsfonts}%
\usepackage{amsthm}%
\usepackage{bm}
\usepackage{mathrsfs}%
\usepackage[title]{appendix}%
\usepackage{xcolor}%
\usepackage{textcomp}%
\usepackage{manyfoot}%
\usepackage{booktabs}%
\usepackage{algorithm}%
\usepackage{algorithmicx}%
\usepackage{algpseudocode}%
\usepackage{listings}%
\usepackage{xspace}
\usepackage{booktabs}
\usepackage{tabulary}
\usepackage{enumitem}
\usepackage{subfig}
\usepackage{pifont}
\usepackage{lmodern}

\usepackage{pdflscape}

\newcolumntype{L}[1]{>{\raggedright\arraybackslash}p{#1}}
\newcolumntype{R}[1]{>{\raggedleft\arraybackslash}p{#1}}
\newcolumntype{C}[1]{>{\centering\arraybackslash}p{#1}}
\newcommand{\ra}[1]{\renewcommand{\arraystretch}{#1}}

\newcommand{\cmark}{\ding{51}}%
\newcommand{\xmark}{\ding{55}}%

\makeatletter
\DeclareRobustCommand\onedot{\futurelet\@let@token\@onedot}
\def\@onedot{\ifx\@let@token.\else.\null\fi\xspace}
\def\eg{\emph{e.g}\onedot} 
\def\ie{\emph{i.e}\onedot}

\def\wrt{w.r.t\onedot} 
\def\etal{\emph{et al}\onedot}

\newlist{inlist}{enumerate*}{1}
\setlist[inlist]{label=(\arabic*)}
\newlist{inlistalpha}{enumerate*}{1}
\setlist[inlistalpha]{label=(\alph*)}

\newcommand{\R}{\ensuremath{\mathbb{R}}}

\begin{document}

\title{Towards Virtual Clinical Trials of Radiology AI with Conditional Generative Modeling}


\author*[1,2,3]{\fnm{Benjamin~D.} \sur{Killeen}}\email{killeen@jhu.edu}
\equalcont{These authors contributed equally to this work.}
\author[1]{\fnm{Bohua} \sur{Wan}}\email{bwan2@jhu.edu}
\equalcont{These authors contributed equally to this work.}

\author[2,4]{\fnm{Aditya~V.} \sur{Kulkarni}}\email{akulka11@jhu.edu}
\author[1,5]{\fnm{Nathan} \sur{Drenkow}}\email{ndrenko1@jhu.edu}
\author[1,3]{\fnm{Michael} \sur{Oberst}}\email{moberst@jhu.edu}
\author[3,4]{\fnm{Paul~H.} \sur{Yi}}\email{paul.yi@stjude.org}
\author*[1,2,3]{\fnm{Mathias} \sur{Unberath}}\email{unberath@jhu.edu}

\affil*[1]{\orgdiv{Department of Computer Science}, \orgname{Johns Hopkins University}, \orgaddress{\city{Baltimore}, \postcode{21218}, \state{MD}, \country{USA}}}

\affil*[2]{\orgdiv{Laboratory for Computational Sensing and Robotics}, \orgname{Johns Hopkins University}, \orgaddress{\city{Baltimore}, \postcode{21218}, \state{MD}, \country{USA}}}

\affil*[3]{\orgdiv{Malone Center for Healthcare in Engineering}, \orgname{Johns Hopkins University}, \orgaddress{\city{Baltimore}, \postcode{21218}, \state{MD}, \country{USA}}}

\affil*[4]{\orgdiv{Department of Radiology}, \orgname{St. Jude Children’s Research Hospital}, \orgaddress{\street{262 Danny Thomas Place}, \city{Memphis}, \postcode{38105-3678}, \state{TN}, \country{USA}}}

\affil*[5]{\orgdiv{Applied Physics Laboratory}, \orgname{Johns Hopkins University}, \orgaddress{\city{Laurel}, \postcode{20723}, \state{MD}, \country{USA}}}



\abstract{
  Artificial intelligence and machine learning (AI/ML) are poised to transform healthcare by enabling personalized and efficient patient care through data-driven insights. Although radiology is at the forefront of AI/ML adoption, in practice, the potential of AI/ML models is often overshadowed by severe failures to generalize: AI/ML models can have performance degradation of up to 20\% when transitioning from controlled test environments to clinical use by radiologists. This mismatch in advertised and observed AI/ML performance raises concerns that radiologists will be misled by incorrect AI/ML predictions in practice and/or grow to distrust AI/ML, rendering these promising technologies practically ineffectual. 
  Exhaustive clinical trials of AI/ML models throughout the development cycle on abundant and diverse data is thus critical to anticipate AI/ML model degradation when encountering varied data samples. Achieving these goals in practice, however, is challenging due to the high costs of collecting the necessary diverse data samples and the corresponding annotations.
  To overcome these limitations, we introduce a novel conditional generative AI model designed for virtual clinical trials (VCTs) of radiology AI/ML, capable of realistically synthesizing full-body CT images of patients with specified attributes. 
  By learning the joint distribution of images and anatomical structures, and operating on latent representations for memory efficiency, our model enables precise replication of real-world patient populations with unprecedented detail at this scale.
  We demonstrate meaningful evaluation of radiology AI models through VCTs powered by our synthetic CT study populations, revealing model degradation and facilitating algorithmic auditing for bias-inducing data attributes.
  Our generative AI approach to VCTs is a promising avenue towards a scalable solution to assess model robustness, mitigate biases, and safeguard patient care by enabling simpler testing and evaluation of AI/ML models in any desired range of diverse patient populations.
}




\maketitle


\section{Main}
\label{sec:intro}

\begin{figure}[t]
  \centering
  \includegraphics[width=\linewidth]{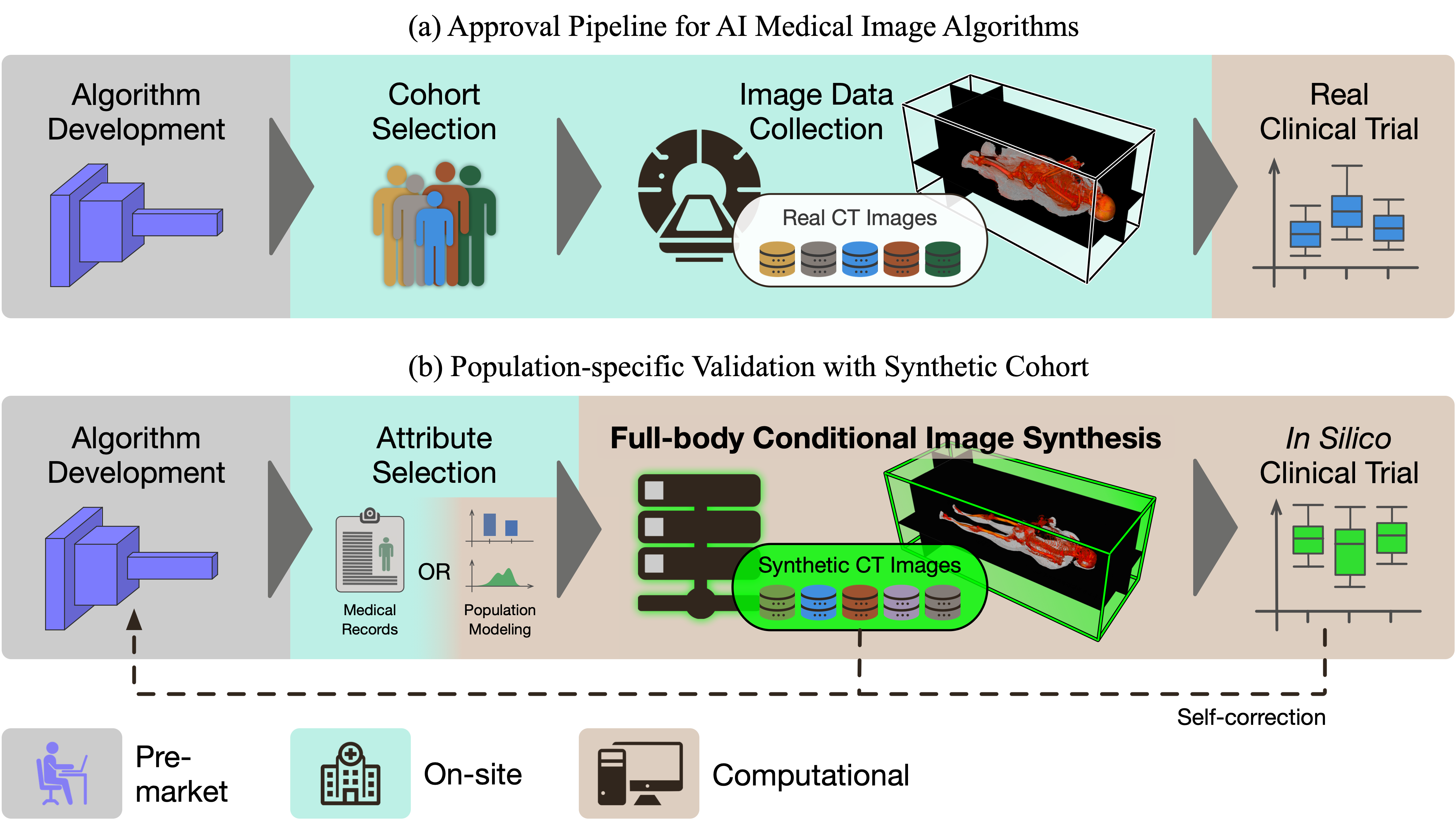}
  \caption{AI-based medical image analysis algorithms are susceptible to drops in performance when deployed on new populations. (a) The approval pipeline for medical image AI necessitates large cohort selection and costly data collection processes so as to ensure good performance across the given population. Performance may still decline when deployed on new populations.\cite{oakden2020hidden}
    (b) We propose a novel framework for medical image AI validation, where a conditional generative model provides full-body images with the same distribution of attributes, \ie. demographics or other characteristics, as the target population. This enables \emph{in silico} clinical trials much earlier in the development pipeline, ensuring high performance on desired populations before real clinical trials.
  }
  \label{fig:overview}
\end{figure}

Artificial intelligence and machine learning (AI/ML) have the potential to transform healthcare by deriving actionable insights into personalized care from vast amounts of data (\ie, precision medicine).
The opportunities to improve the quality, efficiency, and accessibility of healthcare through AI/ML are numerous, with radiology and particularly CT image analysis being being a prime example.
Potential applications in this area include triage acceleration,\cite{titano2018automated, levin2018machine} disease and injury detection,\cite{liu2019comparison, pickhardt2013opportunistic, jang2019opportunistic,eng2021automated, oh2024evaluation} body composition measurement,\cite{zopfs2020evaluating} and clinical decision-making.\cite{ozsahin2020review, murugesan2022a, mohammadi2024deep}
In light of the potential benefits of AI/ML technologies, FDA clearances for AI/ML-based ``software as a medical device (SaMD)'' have surged from 29 in 2020\cite{benjamens2020state} to over 1016 in 2024,\cite{fda2024artificial} with a considerable portion aimed at medical image analysis for radiology.\cite{benjamens2020state}
In many cases, a key requirement of regulatory clearance is demonstration of robust performance in controlled trials, as in Fig.~\ref{fig:overview}a. However, substantial evidence points toward the brittleness of AI/ML models for image analysis, where performance often degrades significantly when deployed outside controlled environments.\cite{drenkow2021systematic} In fact, recent studies indicate that controlled trials can overestimate AI/ML performance by 20\% or more,\cite{Ong_Ly2024-jh, drenkow2021systematic, Drenkow2023-qj, rodrigues2022automated, rava2021validation, matsoukas2023ai, wong2021external, voter2021diagnostic, small2021ct, kunst2023real} with a significant portion of evaluations based on retrospective data from a small number of institutions.\cite{wu2021how}
These errors often stem from biases, shortcuts, and other differences between the data used for developing and validating models and the images observed from a target population during deployment.
Proven approaches such as site-specific clinical trials or causal inference-based analytics can provide these insights if sufficient data is available,\cite{lam2022randomized, oakden2020hidden} but ongoing data collection, and its annotation, is costly and impractical especially with deteriorating models that may already negatively impact patient care.
Thus, anticipating model degradation through automatic, easily repeatable processes is paramount to guarantee peak model performance on an ongoing basis\cite{arpa2024} and to prevent erroneous outputs from adversely affecting treatment plans and encoding systemic biases into the mechanisms of precision medicine.\cite{rajpurkar2022ai}

One way of overcoming these practical challenges is through virtual clinical trials (VCTs). The goal of VCTs is to replicate the model performance that would occur in a real target population using synthetic images. In this scenario, because both the data and its associated label are precisely known and specified at generation time, VCTs overcome a primary challenge of approaches that require real data. Although existing methods, like computational phantoms, have made some progress towards VCTs for some medical imaging applications,\cite{segars2008realistic, abadi2020virtual, badano2023stochastic} they do not yet offer a clear path toward image generation with sufficient realism, variability, scalability, and control to model diverse populations with reasonably low costs.
Generative AI models, on the other hand, can consume and produce practically unlimited data.\cite{ibrahim2024generative} They have been used to augment model training with synthetic images,\cite{hung2023med} improving the performance of downstream AI models for radiology.\cite{khader2023denoising}
Generative models are highly flexible in the kind of conditioning parameters they can incorporate,\cite{chen2024medical} including any attributes that may lead to model degradation.
While conditional generative models exist for other modalities, \eg, for chest X-ray,\cite{chen2024medical} no CT image generative model has been developed with the capabilities necessary for conducting VCTs in radiology AI.
First, volumetric full-body images are required as output in order to represent attributes based on an entire patient, such as height and weight, in an easily verifiable format. Second, the generated images must be sufficiently realistic in terms of visual features and anatomical structure, to ensure that observed changes in performance can be attributed to real model degradations and not domain gaps between synthetic and real images.\cite{gao2023synthex} Finally, the model should generate images with high fidelity to conditioning parameters, enabling VCTs to replicate target populations based solely on attributes that can be collected from medical records or modeled based on survey data,\cite{hennessy2015the} as in Fig.~\ref{fig:overview}b.

Here, we present the first CT image generative model to fully embody these capabilities, as demonstrated through multiple VCTs that replicate model performance and anticipate hidden biases for multiple tasks in radiology-based precision medicine.
A significant challenge for our approach is the complexity of human anatomy, especially over the full body.
Previous work has leveraged a strong prior on the anatomical shape, such as an image in a different imaging modality or a detailed organ segmentation $\mathbf{Y}$, to model the conditional distribution $p(\mathbf{X}|\mathbf{Y})$ of CT images $\mathbf{X}$.\cite{guo2024maisi}
Here, we incorporate anatomical consistency into a model by learning the joint distribution $p(\mathbf{X}, \mathbf{Y})$. This results in images that are visually and anatomically accurate, even compared to methods that focus on smaller regions and have access to detailed anatomical structure information through segmentation.
To accomplish this over the full body in a memory-efficient manner, our generative model operates on latent representations of the full-body image and segmentation.\cite{esser2020taming, rombach2022high}
To support VCTs, we further model the distribution $p(\mathbf{X},\mathbf{Y}|\mathbf{a})$ conditioned on patient attributes $\mathbf{a}$ to allow for sampling of synthetic target populations. In our experiments, we model populations based on demographic attributes (\ie, sex, age, height, and weight) that are relevant to tasks in precision medicine.
In controlled experiments, we show how VCTs using these synthetic cohorts can identify areas of bias and performance degradation for downstream models across multiple tasks in radiology AI, without requiring real data that would be otherwise inaccessible.
The \textbf{generative model} is the latent diffusion model used to sample synthetic images, while the AI model being evaluated in the VCT is the \textbf{downstream model}.
In particular, we consider downstream models that estimate quantities like overall body fat and muscle mass percentage, and we recover known failure modes using only synthetic data.  While these models perform well on data from the same attribute distribution as the training and validation data, they fail to generalize across diverse populations.
When we proactively test with a synthetic patient cohort, we're able to predict the same performance degradation and underlying causes as seen in the deployment case.
Together, these capabilities enable a vendor, hospital, or regulatory body to anticipate AI/ML deployment time changes in performance, identify the population attributes responsible for such changes, and, ultimately prevent adverse effects on patient care and perpetuation of biases in medical data that reflect real-world health disparities.

\subsection{A Conditional Generative Model for Full-body CT Synthesis}
\label{sec:model-overview}

\begin{figure}[t]
  \centering
  \includegraphics[width=\linewidth]{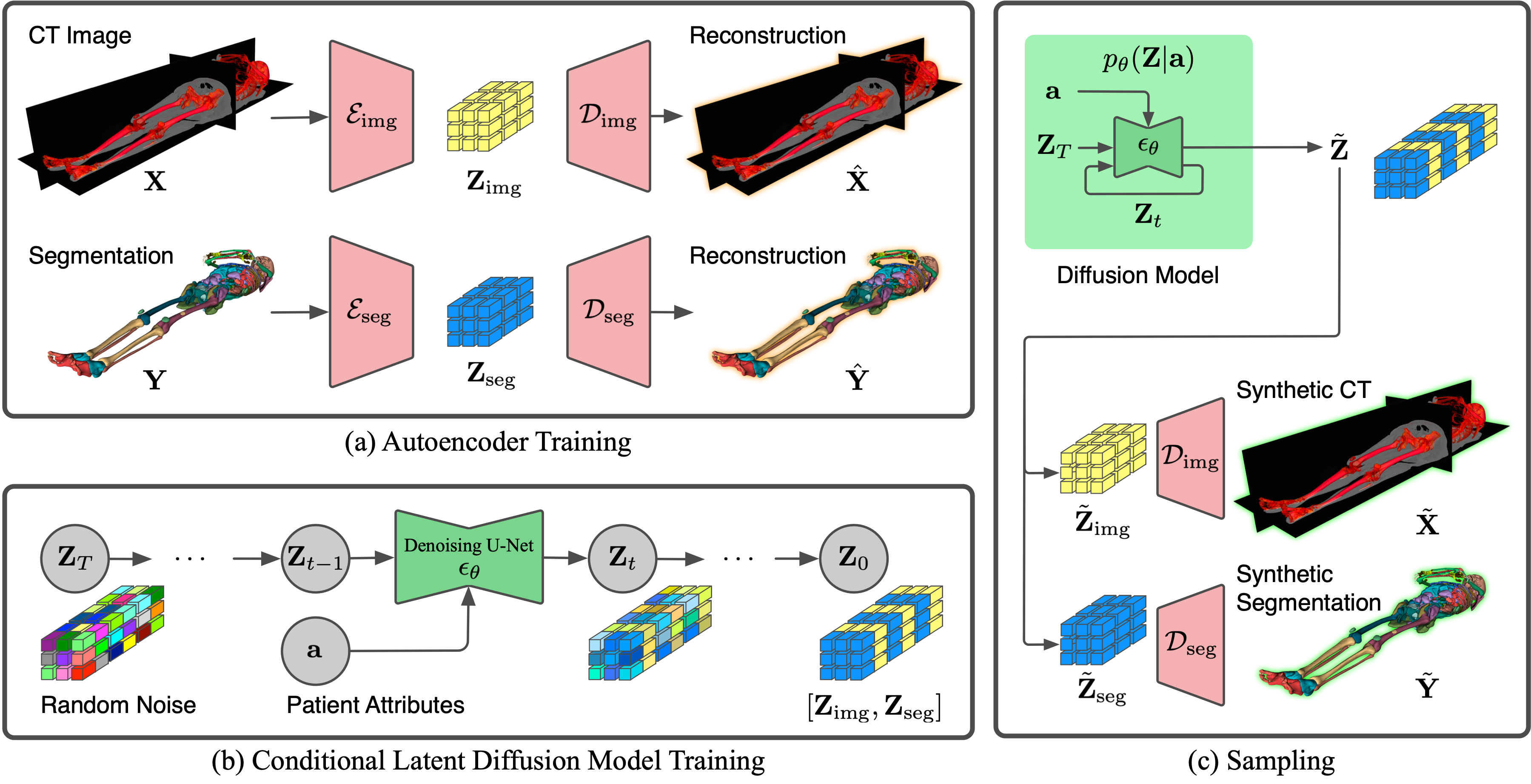}
  \caption{A conditional generative model for full body CT synthesis. (a) Two autoencoders are responsible for compressing the 3D image and segmentation to latent embeddings $\mathbf{Z}_{\rm img}$ and $\mathbf{Z}_{\rm img}$, respectively. (b) A denoising diffusion model learns to sample the distribution for paired embeddings $\mathbf{Z} = [\mathbf{Z}_{\rm img}, \mathbf{Z}_{\rm seg}]$, conditioned on patient attributes $\mathbf{a}$. (c) During image synthesis, the diffusion model samples a random latent code $\mathbf{Z}$, which is decoded separately into the synthetic CT and corresponding segmentation.}
  \label{fig:model-overview}
\end{figure}

Our generative model consists of three main components:
\begin{inlist}
\item an image autoencoder,
\item a segmentation autoencoder, and
\item a latent diffusion model.
\end{inlist} The key capability of this model, which enables it to operate on full-body images with high resolution, is the ability to compress the image and segmentation data into a low-dimensional latent space while still enabling high-quality reconstruction. This compression and reconstruction is achieved in a patch-wise manner using a stacked autoencoder architecture, which allows for a high overall compression rate without sacrificing reconstruction quality.\cite{esser2020taming}
As shown in Fig.~\ref{fig:model-overview}a, the image autoencoder $\mathcal{E}_{\rm img}$ and segmentation autoencoder $\mathcal{E}_{\rm seg}$ compress the full-body CT image $\mathbf{X}$ and segmentation $\mathbf{Y}$ into latent embeddings $\mathbf{Z}_{\rm img}$ and $\mathbf{Z}_{\rm seg}$, respectively. The latent embeddings are then used to reconstruct the image and segmentation using the corresponding decoders $\mathcal{D}_{\rm img}$ and $\mathcal{D}_{\rm seg}$.
The latent diffusion model (Fig.~\ref{fig:model-overview}b) is a probabilistic model for the joint distribution $p(\mathbf{Z}_{\rm img}, \mathbf{Z}_{\rm seg} | \mathbf{a})$ of the latent embeddings, conditioned on patient attributes $\mathbf{a}$. In our experiments, the attributes are demographic categories describing the patient's sex, age, height, and weight, which are relevant to the tasks in precision medicine that we evaluate. During CT synthesis, the diffusion model samples a random latent code $\mathbf{Z} = [\mathbf{Z}_{\rm img}, \mathbf{Z}_{\rm seg}]$ from the learned distribution, which is then decoded into a synthetic image $\tilde{\mathbf{X}}$ and segmentation $\tilde{\mathbf{Y}}$, as in Fig.~\ref{fig:model-overview}c.

\subsection{Evaluation Metrics}
\label{sec:metrics}

In evaluating our generative model and the VCTs it enables, we are often interested in assessing the similarity between univariate distributions. For example, may want to compare the distribution of measured height in the synthetic population to that in the target population, to evaluate the model's fidelity to conditioning parameters. For VCTs, we are primarily interested in whether the absolute error on synthetic images is representative of the absolute error on real images. To quantify the difference between samples from two distributions, we use the standard score (Z-score), which measures the difference in standard deviations. Given two sets of samples $\mathbf{x}$ and $\mathbf{y}$, the Z-score is defined as
\begin{equation}
  \label{eq:zscore}
  Z(\mathbf{x}, \mathbf{y}) = \frac{\bar{x} - \bar{y}}{\sqrt{
      \frac{\sigma_{x}^2}{|\mathbf{x}|} + \frac{\sigma_{y}^2}{|\mathbf{y}|}
  }}.
\end{equation}
A Z-score of 0 indicates that the distributions are identical in terms of mean and variance.
In Section~\ref{sec:downstream}, we obtain 95\% confidence intervals on the Z-scores using bootstrapping, which involves resampling the data with replacement and computing the Z-score for each resample. A narrow interval indicates higher confidence that the distributions are the same.

\subsection{Synthetic Image Realism}
\label{sec:realism}

\begin{figure}
  \centering
  \includegraphics[width=\linewidth]{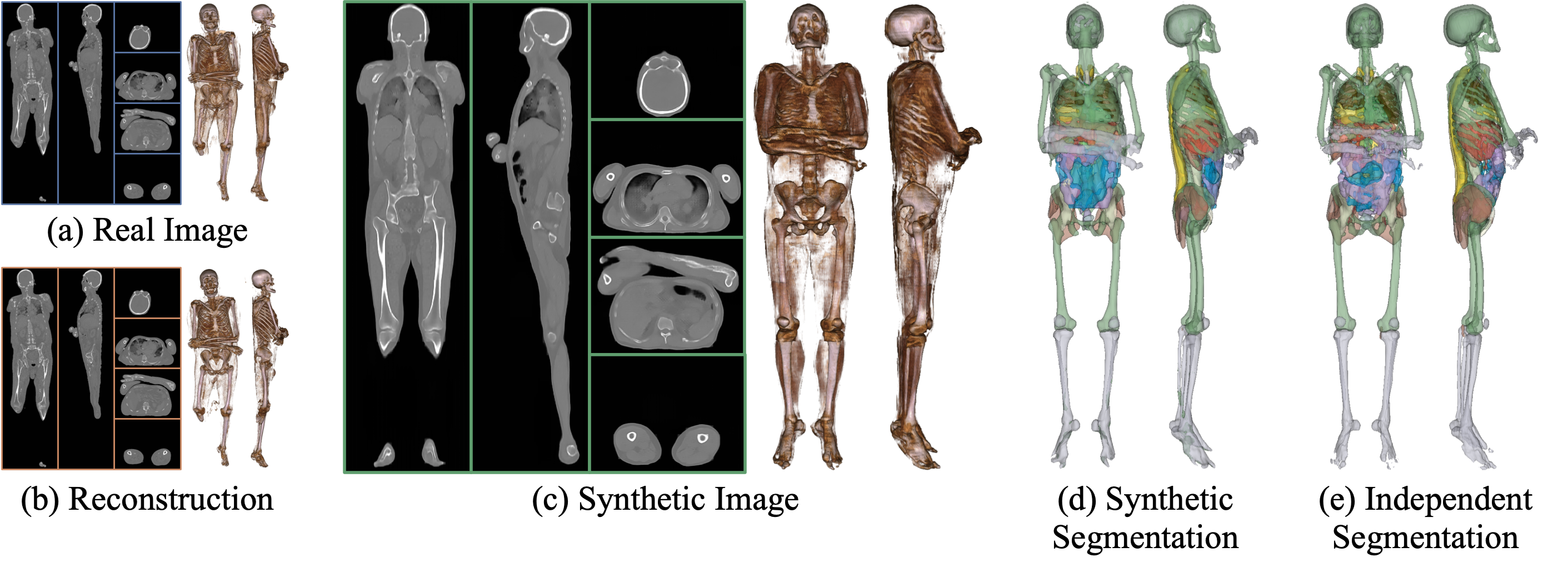}
  \caption{Example outputs from the model. (a) A real image in the training set, in this case from a 66 year old male measuring 180\,cm and 70\,kg, with an amputated right leg. (b) The corresponding VQ-VAE reconstruction of the image. (c) A synthetic sample conditioned to align with the same patient attributes (male, 50-60 years old, 170-180\,cm, and 60 - 70\,kg). Since missing limbs are not included in conditioning, the synthetic image reflects the general population rather than the corresponding case in the training set. (d) The synthetic segmentation generated alongside (c). An independent segmentation of the synthetic image (c) using TotalSegmentator,\cite{wasserthal2023totalsegmentator} with a corresponding class mapping.}
  \label{fig:qualitative-sample}
\end{figure}

The generative model described above is able to synthesize full-body CT images, as in Fig.~\ref{fig:qualitative-sample}, that are realistic in terms of visual features as well as anatomical consistency. This is important for ensuring that there is not a significant domain gap between synthetic and real images, which may cause performance degradation to be observed in VCTs even if the downstream model is robust in real images.\cite{gao2023synthex}
We assess visual realism for reconstructions and synthetic samples primarily using the Frech\'et Inception Distance (FID), a widely used metric for quantifying the similarity between synthetic and real images,\cite{heusel2017gans} including medical images.\cite{asadi2021artificial} Although FID utilizes embeddings from an Inception V3 convolutional neural network that has been pre-trained to classify natural images, it has been shown to effectively evaluate the realism of CT images when using an appropriate dataset for comparison.\cite{oreilly2021pretrained}
The FID of the full-body images when using a stacked image decoder and latent diffusion model for the joint distribution was 5.97, comparable to related work.
Guo \etal\cite{guo2024maisi}, for example, achieve an FID score of 6.083 using the autoPET 2023 dataset as a reference.\cite{gatidis2022autopet}

Beyond low-level visual realism, the anatomical accuracy of synthetic images is important for VCTs in precision medicine as well as many other downstream applications. We first assess the internal consistency of the joint diffusion model $p_{\theta}(\mathbf{Z}_{\rm img}, \mathbf{Z}_{\rm seg}|\mathbf{a})$ by evaluating consistency between the decoded segmentation $\tilde{\mathbf{Y}} = \mathcal{D}_{\rm seg}(\tilde{\mathbf{Z}}_{\rm seg})$ and independent segmentation of $\tilde{\mathbf{X}}$, using TotalSegmentator.\cite{wasserthal2023totalsegmentator}
We find that the Dice similarity coefficient between the two segmentations is 0.727, indicating a high degree of consistency between the images and segmentations synthesized by the joint distribution model. Assuming that $\tilde{\mathbf{Y}}$ is a plausible representation of human anatomy, this shows the synthetic image $\tilde{\mathbf{X}}$ shares that realism. To verify this assumption, we further compare the distribution of organ volumes in synthetic images to that in real images, using the TotalSegmentator segmentation of each, based on relative position in the patient-specific RAS coordinate system. We find that the distribution in synthetic images is highly similar to that in real images, with an average Pearson coefficient of 0.911 in the volume of organs and 0.956 in the organ centroids. These findings, which are summarized in Table~\ref{tab:organ-seg}, indicate that the generative model synthesizes full-body CT images with plausible organ sizes and positions for the included classes.

\subsection{Fidelity to Conditioning}
\label{sec:alignment}

\begin{figure}[ht!]
  \centering
  \includegraphics[width=0.9\linewidth]{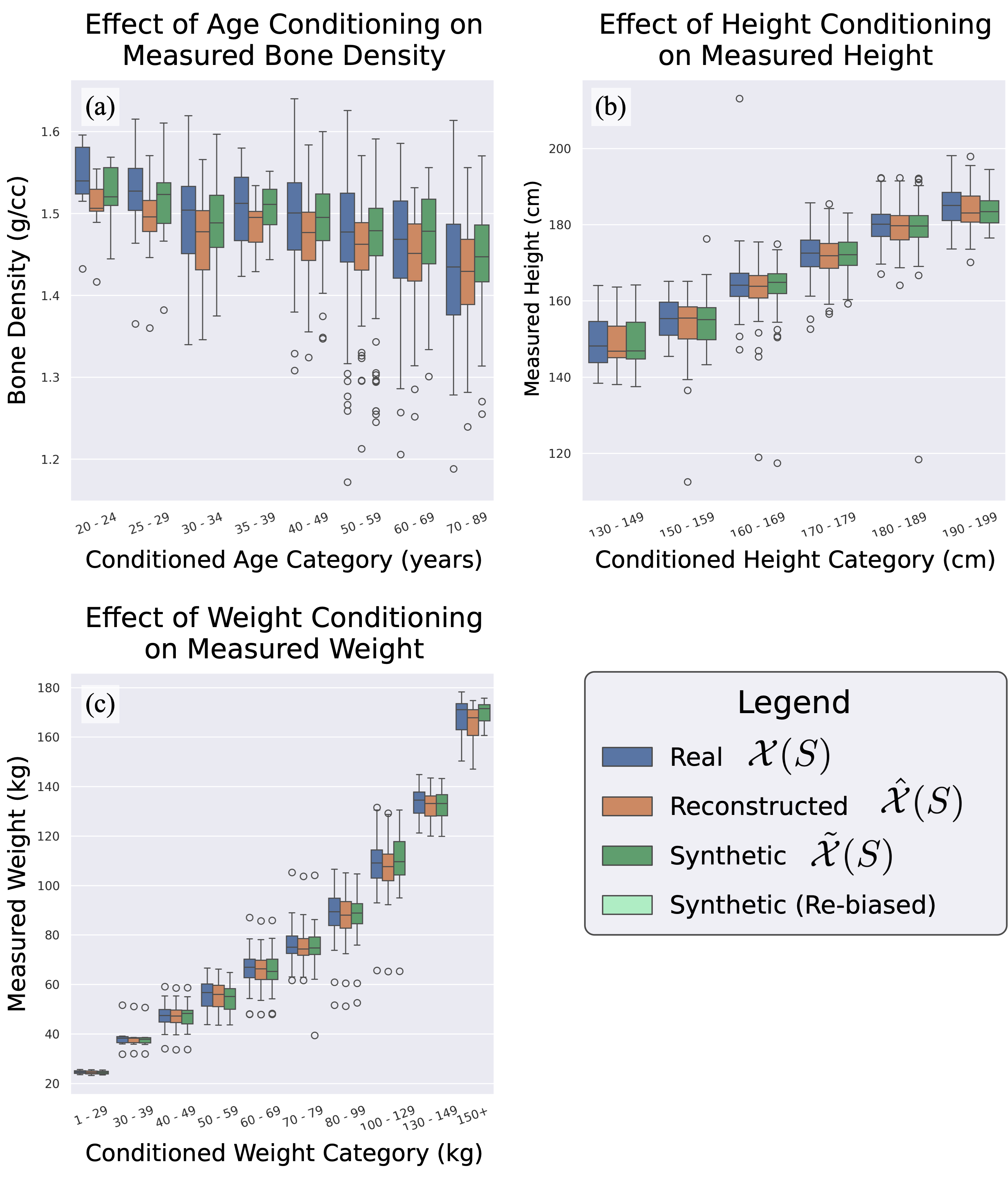}
  \caption{Our model's fidelity to the conditioning categories for age, height, and weight. We show the distribution of measured values based on real CT images, the same images reconstructed with the VQ-VAE, and synthetic images sampled from the same conditioning attributes. The boxes show the quartiles, with whiskers extending to include all inliers. Outliers, as determined based on the inter-quartile range, are shown independently. Because all measurements are calculated using the CT image and an independent organ segmentation,\cite{wasserthal2023totalsegmentator} the conditioned and measured attribute may differ, even for real images. Nevertheless, the alignment between the measured values in synthetic and real images shows our generative model's conditioning faithfully reflects the relevant properties in the real data.}
  \label{fig:quantitative-alignment}
\end{figure}

We evaluate our model's fidelity to conditioning by independently assessing the relevant values from each CT image and comparing them to the conditioned attribute category.
For biological sex, we manually inspect 100 synthetic images randomly sampled with male or female conditioning, finding that sex conditioning results in the correct anatomy in 98\% of cases.
For age, height, and weight, we measure the relevant attribute from the CT image alone, using an independent organ segmentation.\cite{wasserthal2023totalsegmentator} For this experiment, we sample synthetic images conditioned on the same attributes as each real image, with a one-to-one correspondence, so as to ensure realistic combinations of attributes.
Because this measurement may differ systematically from the clinically measured value used for conditioning (see Section~\ref{sec:body-measures}), we make the same measurement across real, reconstructed, and synthetic images, comparing the distribution of measured values.
For age, we examine the average bone density of the images, which is correlated with age.\cite{leichter1981effect} As shown in Fig.~\ref{fig:quantitative-alignment}a, the distribution of bone density values measured for each age conditioning category closely align with real images, with an average Z-score of 0.608 standard deviations across all age categories. For height and weight, we directly measure the conditioned attribute from the CT image, as in Fig.~\ref{fig:quantitative-alignment}b and c. The distributions of measured values for synthetic images closely overlap with those of real images, with average Z-scores of 0.630 and 0.656 for weight and height, respectively.

\subsection{Virtual Clinical Trials for Radiology AI}
\label{sec:downstream}


In this section, we show that VCTs using synthetic images can replicate model performance and identify biases in downstream models for radiology AI.
We focus on two tasks in precision medicine, body fat percentage (BFP) regression and muscle mass percentage (MMP) regression, which are important capabilities for opportunistic body composition measurement.\cite{bates2022ct} To obtain ground truth, we use automated segmentations of tissue types\cite{isensee2021nnu, wasserthal2023totalsegmentator} to compute the mass ratio between the tissue type and the full body (see Section~\ref{sec:body-measures}). This also allows us to compute ground truth for synthetic images, in order to determine the downstream model error.
For the downstream model, we use a deep neural network (DNN) to regress the target variable from 2D coronal and sagittal slices of the input image. Many clinical scenarios favor this ``2.5D'' approach, in which the model takes in multiple 2D slices that together capture 3D information about the patient, because it is significantly less computationally expensive than fully 3D models.\cite{avesta2023comparing} In the context of body composition measurement, an error of about 2 percentage points or less is considered acceptable, while an average error above 3 percentage points is considered significant.\cite{stevens2016prediction,rathnayake2020development,talma2013bioelectrical}
To put this in context, American males have an average BFP from 22.9\% at 16-19 years old to 30.9\% at 60 - 79, as of 2009. Females range from 32.05\% to 42.4\%, based on dual-energy X-ray absorptiometry scans.\cite{cdc2009quickstats}
MMP has been measured using full-body magnetic resonance imaging (MRI) at $38.4 \pm 5.1$\% for males and $30.6 \pm 5.5$\% for females.\cite{janssen2000skeletal}
Note that we report the absolute error in terms of BFP or MMP; although the units are percentage points, these are absolute differences in the percentage of the original body mass, not percentage of the regressed quantity.

To highlight how VCTs using synthetic CT images can detect model degradation, we intentionally sample a biased training set with a \emph{shortcut}---that is, an easily detectable feature that is correlated with the output variable despite being non-clinically relevant.\cite{pavlak2023data} We bias the training set to have a high correlation between the body volume and the target variable, \ie, body fat or muscle mass percentage. We divide the withheld test set into two populations, an in-distribution (ID) population with the same bias as the training set, and an out-of-distribution (OOD) population with a different bias. For example, an ID population with high correlation between body volume and BFP will facilitate shortcut learning based on a specific linear relationship, but the corresponding OOD population will feature a different linear relationship between body volume and BFP.
Overall, this replicates the real-world scenario where real images and patient attributes are available for training and validation on a given distribution, but only demographic attributes are available from a target population (\eg, a hospital population where a model is to be deployed).
Having access to the real OOD images in our experiments allows us to compare the performance of each downstream model on synthetic images with the real ones, isolating whether the observed degradation is due to the bias in the training data.
In this context, we are interested in two questions. First, does a VCT detect model degradation from the ID data to the OOD data, based on synthetic images with the same distribution of attributes? Second, do VCTs with synthetic images reveal the exact kind of bias in the downstream model's performance on real images? This section examines these questions.

\subsubsection{Detecting Model Degradation via VCTs}
\label{sec:vct-degradation}
\noindent\textbf{\textit{Does a VCT detect model degradation from the ID data to the
OOD data, based on synthetic images with the same distribution of attributes? }}


For both tasks, we observe significant model degradation on the real OOD population, despite good performance on the ID population.
As shown in Table~\ref{tab:z-scores}, on the real ID test set $f_{\rm BFP}$ achieves a mean absolute error (MAE) of 1.20\% (95\% CI: 1.04 to 1.40\%) and $f_{\rm MMP}$ achieves an MAE of 1.43\% (95\% CI: 1.18 to 1.73\%), confidently within a nominally acceptable error of 2 percentage points.
This indicates that both models are capable of accurately estimating the target variable from CT slices and, in our scenario, may obtain regulatory clearance based on ID performance.
However, on the real OOD test set, both $f_{\rm BFP}$ and $f_{\rm MMP}$ degrade to unacceptable error levels, achieving an MAE of 3.66\% (95\% CI: 3.10 to 4.31\%) and 5.54\% (95\% CI: 5.02 to 6.16\%), respectively.
This indicates the models are not robust to the population shift, and may lead to adverse effects on patient care if deployed without further validation.

Conventional approaches to anticipate model degradation may fail to detect this bias. With the same information available, a straightforward baseline approach is to reweight the errors measured on the ID test set based on the likelihood $p(\text{OOD} | \mathbf{a})$ of coming from the OOD population, so that the model's degradation on the most relevant samples is amplified. This yields an estimated MAE on the real OOD set of 1.31\% (95\% CI: 1.04, 1.54) for BFP, which does not indicate the true MAE on the OOD set of 3.66\% or signify errors outside the acceptable range. For MMP, the weighted MAE is 1.65\% (95\% CI: 1.25, 2.40), which is far from the true value of 5.54\%. These results indicate that conventional statistical approaches for detecting model degradation are not sufficient to detect the bias due to population shift in our experiments.

\begin{figure}[t]
  \centering
  \includegraphics[width=\linewidth]{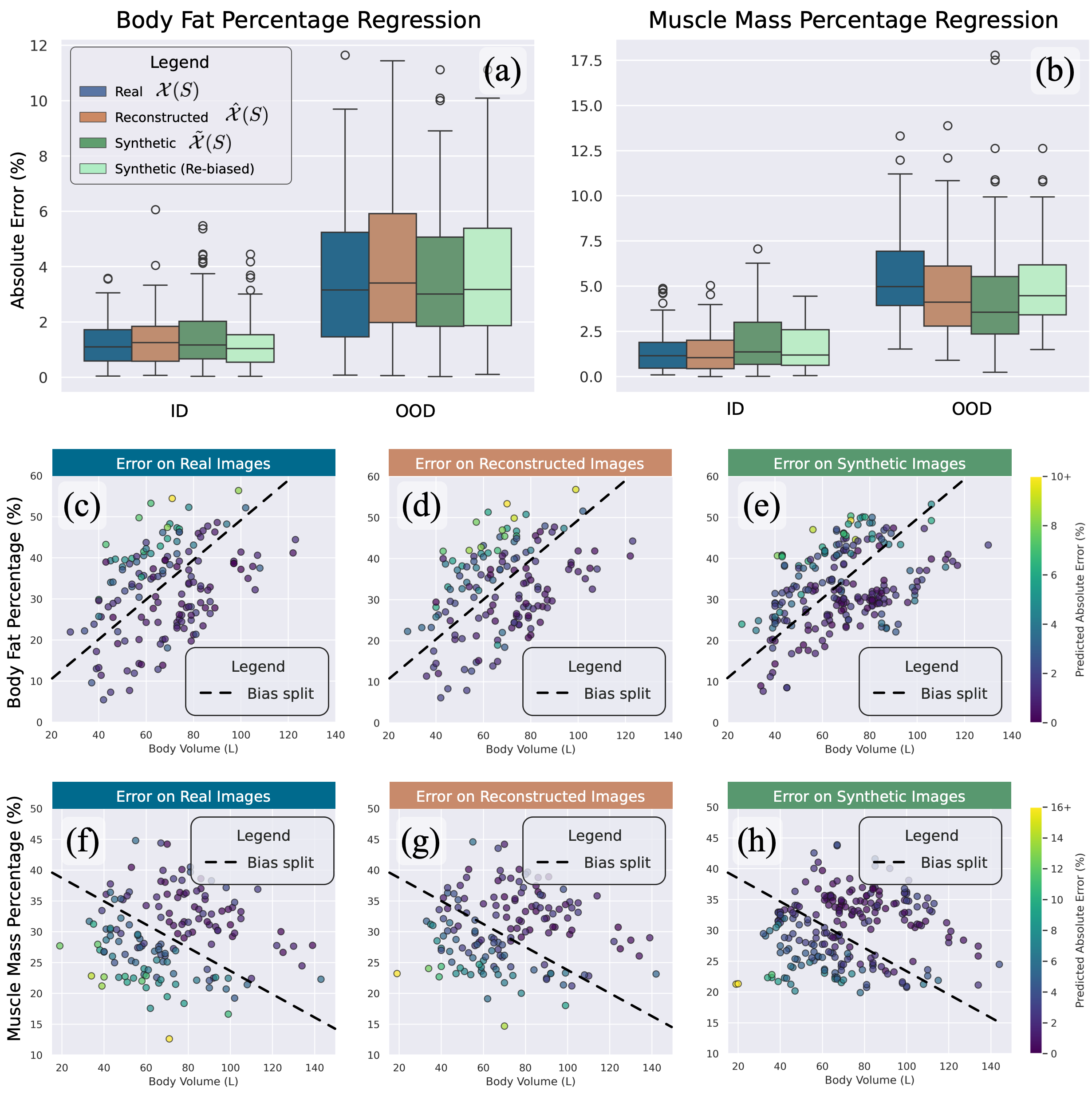}
  \caption{Results of the VCT, including absolute error for BFP and MMP. See Table~\ref{tab:z-scores} for complete quantitative results.
  }
  \label{fig:downstream-vct}
\end{figure}

In contrast, VCTs using synthetic images can detect model degradation in both tasks, based on the distribution of attributes. Because the patient attributes are not fully predictive of ID/OOD status, we oversample the ID and OOD populations by a factor of 2, resulting in two distinct synthetic images conditioned on $\mathbf{a}_i$ for each patient $i$ in the test sets.
For both tasks, we find that the MAE on synthetic images aligns with that of real images, indicating acceptable errors ($< 2\%$) for the ID population and significant errors for the OOD population ($> 3\%$).
There is, however, a difference in the distribution of errors on synthetic and real images.
We hypothesize that this difference arises from the fact that the same patient attributes can result in different body compositions, which would be separated into one population or the other in the real data, but are not separated in the synthetic data.
To test this hypothesis, we re-bias the synthetic data in the same manner as the real images, by culling synthetic images outside the specified distribution.
This results in a close match between the distribution of absolute errors on synthetic and real images, across both the ID and OOD data in both tasks.
Quantitatively, the real MAE falls within the 95\% CI of the synthetic MAE, and a Z-score of 0 standard deviations is in the CI, based on bootstrapping analysis. The Z-test in all cases indicates high probability that the absolute errors are from the same distribution ($p > 0.3$).
Fig.~\ref{fig:downstream-vct}a-b show the full distribution of absolute errors for each task and test set.
For completeness, we also examine the distribution of errors using reconstructed images with a one-to-one correspondence to the real images. The close match between errors on reconstructed images and real images further suggests that the observed differences for synthetic images in the distribution is due to variation in sampling, rather than a significant domain gap between synthetic and real images.
Thus, VCTs using our generative model are capable of reproducing model performance on populations based on patient attributes, as long as the conditioning attributes are sufficient to reproduce the biasing attributes.

  \begin{sidewaystable}[h!]
    \ra{1.3}
    \centering
    \caption{AI model performance on biased populations. For each task, the ID test set comes from the same distribution as the training set, while the out-of-distribution (OOD) is biased with respect to the target variable and body volume. For each population, the goal is to predict model success or failure, as signified by an MAE below 2\% or above 3\%, respectively. A secondary goal is to replicate the distribution of errors on real images of the same population, for which the Z-score and $p$-value are computed with respect to the corresponding error on real images (ideally $\sim 0$ and $>0.05$, respectively). 95\% CIs are based on bootstrapping. As can be seen, additional re-biasing of the synthetic images yields a matching absolute error distribution to real images.}
    \begin{tabular}{@{}p{0.1em} lll r rr rr r}
      \toprule
      \multicolumn{4}{@{}l}{\textbf{Downstream Task}}
      && \multicolumn{2}{c}{MAE} & \multicolumn{2}{c}{Z-Score \wrt Real} & \multicolumn{1}{c}{$p$-value}\\
      \cmidrule(lr){6-7} \cmidrule(lr){8-9} 
      & Population & Attr. Dist. & Sample Type & \# & \multicolumn{1}{r}{Mean} & \multicolumn{1}{r}{95\% CI} &\multicolumn{1}{r}{Mean} & \multicolumn{1}{r}{95\% CI} & \\
      \midrule
      \multicolumn{8}{@{}l}{\textbf{Body Fat Percentage (\%)}} \\ 
      & ID & ID & Real & 75 & 1.20 & [1.04, 1.40] && \multicolumn{1}{c}{---} & \multicolumn{1}{c}{---}\\
      &   & ID & Reconstructed & 75 & 1.34 & [1.15, 1.60] & 0.97 & [-1.02, 2.87] & 0.368 \\
      &   & ID & Synthetic & 150 & 1.50 & [1.31, 1.72] & 2.33 & [0.14, 4.35] & 0.103 \\
      &   & ID & Synthetic (Re-biased) & 91 & 1.22 & [1.05, 1.43] & 0.10 & [-1.99, 2.14] & 0.481 \\
      \cmidrule(lr){2-2}
      & OOD & OOD & Real & 75 & 3.66 & [3.10, 4.31] && \multicolumn{1}{c}{---} & \multicolumn{1}{c}{---}\\
      & & ID & Real (Weighted) & 75 & 1.31 & [1.04, 1.54] & & \multicolumn{1}{c}{---} & \multicolumn{1}{c}{---} \\
      &   & OOD & Reconstructed & 75 & 4.07 & [3.47, 4.76] & 0.90 & [-1.09, 2.88] & 0.387 \\
      &   & OOD & Synthetic & 150 & 3.69 & [3.26, 4.17] & 0.08 & [-2.18, 2.27] & 0.461 \\
      &   & OOD & Synthetic (Re-biased) & 94 & 3.86 & [3.39, 4.39] & 0.53 & [-1.61, 2.63] & 0.439 \\
      \multicolumn{8}{@{}l}{\textbf{Muscle Percentage (\%)}} \\ 
      & ID & ID & Real & 75 & 1.43 & [1.18, 1.73] && \multicolumn{1}{c}{---} & \multicolumn{1}{c}{---}\\
      &   & ID & Reconstructed & 75 & 1.36 & [1.13, 1.65] & -0.33 & [-2.30, 1.67] & 0.480 \\
      &   & ID & Synthetic & 150 & 1.92 & [1.66, 2.22] & 2.74 & [0.45, 4.86] & 0.063 \\
      &   & ID & Synthetic (Re-biased) & 94 & 1.58 & [1.35, 1.85] & 0.90 & [-1.25, 3.01] & 0.379 \\
      \cmidrule(lr){2-2}
      & OOD & OOD & Real & 75 & 5.54 & [5.02, 6.16] && \multicolumn{1}{c}{---} & \multicolumn{1}{c}{---}\\
      & & ID & Real (Weighted) & 75 & 1.65 & [1.25, 2.40] & & \multicolumn{1}{c}{---} & \multicolumn{1}{c}{---} \\
      &   & OOD & Reconstructed & 75 & 4.58 & [4.07, 5.21] & -2.41 & [-4.36, -0.27] & 0.090 \\
      &   & OOD & Synthetic & 150 & 4.34 & [3.86, 4.95] & -3.44 & [-5.70, -0.94] & 0.026 \\
      &   & OOD & Synthetic (Re-biased) & 80 & 5.10 & [4.64, 5.65] & -1.17 & [-3.16, 0.88] & 0.324 \\
      \bottomrule
    \end{tabular}
    \label{tab:z-scores}
  \end{sidewaystable}

\subsubsection{Replicating Model Biases in VCTs}
\label{sec:replicating-bias}
\noindent\textbf{\textit{Do VCTs with synthetic images reveal the exact kind of
bias in the downstream model’s performance on real images? }}

To answer this question, we examine the patient attributes that may be to blame for model error across the combined ID and OOD test sets.
Fig.~\ref{fig:downstream-vct}c-h shows the distribution of the errors with respect to the patient attributes used to bias the training data. Qualitatively, the distribution of attributes and errors in the reconstructed test sets closely matches the real test sets. Likewise the synthetic images show a similar distribution of errors even without corresponding samples, with higher error on the OOD side of the bias split at the furthest points from the boundary.
Quantitatively, Table~\ref{tab:bias-corr} details the Pearson correlation coefficient between each variable and the model error on real and synthetic data, showing close alignment with a Z-test $p$-value indicating high probability of being sampled from the same distribution ($p > 0.3$).
This indicates that the synthetic data replicates the same bias found in the real data with respect to the known biased attributes, which is only possible in this case because the bias is artificially constructed.

To quantify the bias of $f_{\rm BFP}$ and $f_{MMP}$ more broadly, we conduct a feature importance analysis to determine the patient attributes which are most predictive of model degradation.
For each task and image type (real, reconstructed, synthetic), we train a random forest regression model to predict the absolute error of the downstream model based on 8 patient attributes: sex, age, height, weight, body fat percentage, bone density, muscle mass percentage, and body volume.
These features are sufficient to predict the model error with an average MAE of 0.69 percentage points across all image types (see Table~\ref{tab:error-error}).
Feature importance analysis reveals the patient attributes that are most predictive of the model error. We find that for real, reconstructed, and synthetic test sets, the feature importance values are highly correlated. The feature importances for reconstructed and real images were almost perfectly correlated, with a value of 0.998 and 0.991 for $f_{\rm BFP}f$ and $f_{\rm MMP}$, respectively.
Between synthetic and real images, the features importances had a correlation of 0.993 and 0.919, for $f_{\rm BFP}f$ and $f_{\rm MMP}$ respectively, indicating that VCTs are a viable pathway toward identifying the biases that contribute to model degradation.

\begin{table}[t]
  \centering
  \begin{tabular}{@{}p{0.1em}lrrr}
    \toprule
    \multicolumn{2}{@{}l}{\textbf{Downstream Task}}
    & \multicolumn{2}{c}{Correlation with Model Error} & $p$-value\\
    \cmidrule{3-4}
    & Attribute & Real & Synthetic \\
    \midrule
    \multicolumn{4}{@{}l}{\textbf{Body Fat Percentage}} \\
    & Body volume & -0.190 & -0.117 & 0.481\\
    & Body fat percentage & 0.482 & 0.483 & 0.989 \\
    \multicolumn{4}{@{}l}{\textbf{Muscle Mass Percentage}} \\
    & Body volume & -0.469 & -0.407 & 0.468\\
    & Body fat percentage & 0.114 & 0.012 & 0.333 \\
    \bottomrule
  \end{tabular}
  \caption{Correlation of model error on synthetic and real images with bias attributes.}
  \label{tab:bias-corr}
\end{table}

\begin{table}[t]
  \centering
  \caption{Model Error Random Forest Regression Error}
  \begin{tabular}{lrrr}
    \toprule
    Downstream Task & \multicolumn{3}{c}{MAE} \\
    \cmidrule{2-4}
    & Real & Reconstructed & Synthetic \\
    \midrule
    Body fat percentage & 0.72 $\pm$ 0.59 & 0.71 $\pm$ 0.63 &  0.61 $\pm$ 0.55\\
    Muscle mass percentage & 0.68 $\pm$ 0.55 & 0.68 $\pm$ 0.55 & 0.73 $\pm$ 0.57\\
  \end{tabular}
  \label{tab:error-error}
\end{table}

\begin{table}[t]
  \ra{1.3}
  \centering
  \caption{Feature Importance}
  \begin{tabular}{@{}p{0.1em}lrrr}
    \toprule
    \multicolumn{2}{@{}l}{\textbf{Downstream Task}} & \multicolumn{3}{c}{Feature Importance}\\
    \cmidrule{3-5}
    & Attribute & Real & Reconstructed & Synthetic \\
    \midrule
    \multicolumn{4}{@{}l}{\textbf{Body Fat Percentage (\%)}} \\
    & Sex & 0.001 & 0.001 & 0.055 \\
    & Age & 0.040 & 0.046 & 0.031 \\
    & Height & 0.062 & 0.046 & 0.036 \\
    & Weight & 0.045 & 0.086 & 0.057 \\
    & Body fat percentage & 0.033 & 0.040 & 0.040 \\
    & Bone density & 0.023 & 0.041 & 0.076 \\
    & Muscle mass percentage & 0.763 & 0.700 & 0.652 \\
    & Body volume & 0.033 & 0.041 & 0.055 \\
    \cmidrule{3-5}
    & Correlation & \multicolumn{1}{c}{---}& 0.998 & 0.993\\
    \multicolumn{4}{@{}l}{\textbf{Muscle Percentage (\%)}} \\
    & Sex & 0.006 & 0.003 & 0.001 \\
    & Age & 0.010 & 0.020 & 0.020 \\
    & Height & 0.068 & 0.028 & 0.027 \\
    & Weight & 0.180 & 0.226 & 0.324 \\
    & Body fat percentage & 0.081 & 0.059 & 0.054 \\
    & Bone density & 0.024 & 0.023 & 0.033 \\
    & Muscle mass percentage & 0.506 & 0.516 & 0.424 \\
    & Body volume & 0.124 & 0.124 & 0.118 \\
    \cmidrule{3-5}
    & Correlation & \multicolumn{1}{c}{---}& 0.991 & 0.919\\
    \bottomrule
  \end{tabular}
  \label{tab:feat-import}
\end{table}

\section{Discussion}
\label{sec:discussion}

VCTs are a key component in the emerging landscape of AI/ML models for radiology. Our first-of-its-kind generative model demonstrates a scalable, flexible, and highly realistic approach to synthesizing virtual patient cohorts suitable for VCTs in precision medicine. We have shown that our model is capable of synthesizing full-body CT images with a high level of realism in terms of visual appearance and anatomical structure. It can generate images from patient attributes (sex, age, height, and weight) that are readily available from medical records and lend themselves to distribution modeling for VCTs. In a simulated VCT, we demonstrated that validation with these images can replicate real biases in downstream AI models across multiple tasks.

Full-body volumetric image synthesis presents significant challenges, which have confronted in this work.
3D convolutional models are memory intensive, but the need for global consistency in anatomical structures requires a 3D approach.
Even using latent image diffusion,\cite{rombach2022high} image encoders and decoders processing full body volumetric CT images are too large to fit on a single GPU.
Prior work has enabled partial CT image synthesis by splitting tensors across multiple GPUs.\cite{guo2024maisi}
Our approach takes the more traditional patch-wise encoder-decoder strategy with a stacked VQ-VAE\cite{esser2020taming} and a final post-processing network to refine synthetic images.
To further improve image realism, we introduced a novel multi-window loss function that reweights the contributions of soft and hard tissue structures, ensuring that larger gradients from hard tissue structures do not dominate learning.
Our experiments demonstrated the value of this approach in terms of reconstructed and synthetic image quality, achieving an average FID score of 5.97, which is important to reducing the sim-to-real gap for downstream tasks.

In addition to low-level image realism, VCTs require high-level anatomical realism.
Target variables like BFP and MMP are only meaningful for anatomically realistic full-body images for which segmentations of organs and tissue type can be easily obtained.
To achieve global anatomical consistency, our approach included organ segmentations in the latent embedding, using a second autoencoder.
Without this, we found that synthetic images might have low FID but lack basic anatomical structures as evaluated by third-party segmentation models.\cite{wasserthal2023totalsegmentator}
Learning the joint distribution of images and basic organ segmentations yielded generative model with valid organ segmentations closely aligned with real images, in terms of the position and size of segmented organs.

Finally, VCTs require a way to condition image generation on relevant patient attributes.
Our model uses categorical conditioning based on demographic attributes from the available metadata. Independent verification of the sex, age, height, and weight shows successful alignment with the training data in terms of these attributes, although in some cases the measured attribute in the real and synthetic images differed from the value in the metadata. This could be because of variable measuring techniques, such as measuring an individual's weight with their clothes on or measuring the body length with limbs bent.
By assessing the difference between quantities measured in the same way, using TotalSegmentator-derived quantities, we can nevertheless conclude that images sampled with a given attribute will align with real images in terms of that attribute, if not with the nominal value in the decedent record.

Our experiments showed that VCTs using synthetic images were able to detect real model biases with respect to patient attributes.
Downstream models for BFP and MMP regression were trained on biased data, with a shortcut that correlated body volume with the target variable.
This resulted in significant model degradation on real images from the OOD population, which was not detected by conventional approaches.
By generating virtual cohorts of synthetic images with the same distribution of patient attributes, we were able to replicate the model performance on synthetic images, indicating clinically acceptable model performance on ID test data and serious degradation on OOD test data.
Further, we were able to identify the patient attributes that were most predictive of model error, and found that the feature importance analysis was highly correlated between real and synthetic images.
This demonstrates that VCTs can be used to identify the biases that contribute to model degradation, and that the generative model can be used to anticipate real-world biases in downstream models.
In future work, this capability may allow for automatic model adjustment to rectify these biases without additional real data collection or annotation.

Despite this progress, there are some noted limitations with the approach outlined here. First, we make the assumption that the generative AI model is capable of faithfully representing samples from the relevant patient attributes, either because such attributes have been observed during training or because the training was conducted at a scale such that this capability manifests as an emergent property.
While the latter case is a promising direction for future work, it is an open question at what scale such generalization capabilities may emerge.
In our experiments, we demonstrate for the first time that a generative model is capable of replicating the real-world performance of a downstream model for radiology AI applications, assuming that the generative model has been exposed to data with similar attributes as the target population.
Therefore, the generative models that enable VCTs as presented here shift the burden of data collection and annotation from the numerous vendors of AI-based systems to a centralized entity, \eg, a regulatory agency or consortium of institutions.
Increasing the flexibility of conditioning supported by the generative model will further increase the scope of VCTs of the kind presented here.
We have focused on demographic attributes, of the kind generally available in medical records, but more flexible conditioning based on any available medical history would broaden the applicability of VCTs based on generative modeling.
Incorporating any relevant data, from past diagnoses to family history, may require text-based conditioning, although the scale required for such conditioning is much larger.\cite{cho2024medisyn}
Text-based conditioning may also offer an illusion of unlimited conditioning potential when the true distribution of supported patient attributes is much smaller than can be described with natural language.
Nonetheless, increasing the flexibility of conditioning attributes is desirable for another reason, namely to reduce the dependence on independent segmentation tools to provide ground truth data for VCTs using synthetic images.
While the model framework used here, TotalSegmentator, has been widely validated on CT images,\cite{wasserthal2023totalsegmentator}, the approach here is so far limited to VCTs related directly to the conditioning attributes supported or quantities that can be derived from models assumed to be accurate.
While we demonstrate good alignment between the synthetic segmentation and TotalSegmentator, more flexible conditioning would allow for more self-contained VCTs that derive the ground truth from the conditioning signal.

\section{Conclusion}
\label{sec:conclusion}

In conclusion, this work advances the state of generative modeling in precision medicine by introducing a first-of-its-kind conditional generative AI model capable of full-body CT image synthesis for VCTs. By achieving high anatomical and visual realism and precise conditioning on demographic attributes, this model enables scalable, proactive assessments of AI model robustness across diverse populations. Our experiments demonstrate the efficacy of VCTs in detecting performance degradations and biases in medical imaging AI systems, replicating real-world model behavior and identifying the population attributes responsible for degradation. These findings establish a pathway for mitigating biases and safeguarding patient care without the extensive costs and impracticalities of on-going real-world data collection. While the approach highlights the potential of generative AI to revolutionize model validation and robustness assessment, further exploration into broader conditioning capabilities and emergent properties of generative models trained at scale will be crucial. Such advancements could expand the scope of VCTs, enabling a more comprehensive evaluation of AI systems for precision medicine and fostering their safe and equitable deployment.


\newpage

\section{Methods}
\label{sec:methods}

\subsection{A Latent Diffusion Model for Conditional Full-body CT Synthesis}
\label{sec:architecture}


Figure \ref{fig:model-overview} shows the overall structure of our generative model, which is composed of 4 parts: 1. a stacked CT image autoencoder, $(\mathcal{E}_{\rm img}=\mathcal{E}^{(2)}\circ \mathcal{E}^{(1)},\mathcal{D}_{\rm img}=\mathcal{D}^{(1)}\circ \mathcal{D}^{(2)})$, that compress input CT image to latent CT vector, $\mathbf{Z}_{\rm img}$, with high compression ratio while preserving anatomical structures. 2. a segmentation autoencoder, $\{\mathcal{E}_{\rm seg},\mathcal{D}_{\rm seg}\}$, that compress segmentation to latent segmentation vector, $\mathbf{Z}_{\rm seg}$, with the same compression ratio. 3. a latent diffusion model for conditional latent vector sampling. 4. a 3D U-Net based post-processing model that further improves the realism of the generated samples.

\subsubsection{Stacked Autoencoder}
\label{sec:stacked-autoencoder}
We propose a framework for stacking autoencoders to achieve better performance in terms of preserving anatomical structures while compressing images to extreme. Vanilla autoencoder such as Vector Quantized Variational Autoencoder (VQ-VAE) \cite{van2017neural}, and Vector Quantized Generative Adversarial Network (VQ-GAN) \cite{esser2020taming} first compress images to latent vectors with a single encoder and then decompress the latent vectors back to reconstructed images with a single decoder. Although using a pair of single encoder and single decoder is simpler, it limits the reconstruction quality and the compression rate. The latent vectors produced by these models are typically 4 to 8 times smaller than the original images in spatial dimensions (height, width, and depth). It has been shown that the reconstruction quality decreases as the compression ratio increases \cite{esser2020taming}. In this approach, we stack multiple encoders and decoders instead. The compression rate of each pair of encoder and decoder is kept small to reduce the difficulties in learning, as it is considerably harder to train encoder and decoder with high compression rate (16 for example) than to train encoder and decoder with low compression rate. The training of each pair of encoder and decoder is separate, thus the model size of each pair is not limited by the number of levels of stacking and training larger model with limited memory is made possible.

Due to computational limitation, all autoencoders are implemented in a patch-based manner and image-level reconstructions and latent vectors are obtained using sliding window\cite{monai_consortium_2024_13942962} with patch-based autoencoders. In the following text, we use \textbf{bolded} lower case letter to represent a patch of an image and use its upper case letter to denote the whole image. For example, ${\bm x}\in \mathbb{R}^{h\times w\times d}$ denotes a $(h,w,d)$ sized patch of an input image $\mathbf{X}\in \mathbb{R}^{H\times W\times D}$ of size $(H,W,D)$.

Formally, we define a stacked autoencoder as $(\mathcal{E},\mathcal{D})^{\rm (L_{ae})}$, where $\mathcal{E}=\mathcal{E}^{\rm (L_{\rm ae})}\circ \mathcal{E}^{\rm (L_{\rm ae}-1)}\circ ... \circ \mathcal{E}^{(1)}$ and $ \mathcal{D}=\mathcal{(D)}^{(1)}\circ \mathcal{D}^{(2)}\circ ... \circ \mathcal{D}^{\rm (L_{\rm ae})}$. ${\rm L_{\rm ae}}\in \mathbb{Z}$ denotes the maximum level of stacking and $\circ$ denotes composition. The vanilla autoencoder is a special case when ${\rm L_{\rm ae}}=1$.

\begin{align*}
  {\bm x}^{(l-1)}= &
  \begin{cases}
    \mathcal{E}^{(l)}({\bm x}^{(l)}) &  1\leq l<{\rm L_{\rm ae}}\\
    \mathcal{E}^{(l)}({\bm x}) & l={\rm L_{\rm ae}}
  \end{cases}.
\end{align*}
, where ${\bm x}^{(l)}$ is the latent vector of a patch encoded by the encoder from level $l$. The compression rate of a stacked autoencode is the multiplication of compression rates of all its encoders. Let ${\bm z} = \mathcal{E}({\bm x})$ be the latent vector of a patch compressed by the encoders. A 2 level stacked autoencoder is used as our CT autoencoder, $(\mathcal{E}_{\rm img},\mathcal{D}_{\rm img})=(\mathcal{E}, \mathcal{D})^{\rm (2)}$, and $\mathcal{E}_{\rm img}=\mathcal{E}^{(2)}\circ\mathcal{E}^{(1)},\mathcal{D}_{\rm img}=\mathcal{D}^{(1)}\circ\mathcal{D}^{(2)}$.

During training, the pairs of encoder and decoder are trained from higher level to lower level and where the lower level reconstructs the encoded latent vectors from the previous level. Each pair $(\mathcal{E}^{(l)},\mathcal{D}^{(l)})$ is trained to minimize $\mathcal{L}^{(l)}_{\rm rec}({\hat{\bm x}}^{(l)}, {\bm x}^{(l)})$, where ${\hat{\bm x}}^{(l)}=\mathcal{D}^{(l)}(\mathcal{E}^{(l)}({\bm x}^{(l)}))$ is the reconstructed input and $\mathcal{L}^{(l)}_{\rm rec}$ is a reconstruction loss for the pair at level $l$ that characterizes the distance between inputs.
\begin{align}
  \mathcal{L}_{\rm rec}^{(l)} =
  \begin{cases}
    \mathcal{L}_{\rm mw} +\mathcal{L}_{\rm per} + \mathcal{L}_{\rm GAN}  &  l = {\rm L_{\rm ae}} \\
    {\rm L1} + \mathcal{L}_{\rm GAN}   & 1\leq l < {\rm L_{\rm ae}}
  \end{cases}.
\end{align}

In the highest layer where the images are used as input, $\mathcal{L}_{rec}^{\rm (L_{\rm ae})}$ combines perceptual loss\cite{johnson2016perceptual}, a GAN loss\cite{esser2020taming}, and a multi-window ${\rm L1}$ loss.

The multi-window L1 loss is a simple modification to L1 loss that re-scales the image regions so that the soft tissue regions contribute more equally to the loss gradient---compared to hard tissue regions---than it would otherwise. For two scalar voxel values $x, \hat{x} \in \R$, let
\begin{align}
  \label{eq:4}
  \mathcal{L}_{\rm mw}(x, \hat{x}) =
  \begin{cases}
    \lambda_{\rm soft} |x - \hat{x}| & \text{if } \text{HU}_{\rm soft}^{\rm min}\leq x < \text{HU}_{\rm soft}^{\rm max}\\
    \lambda_{\rm hard} |x - \hat{x}| & \text{if } \text{HU}_{\rm hard}^{\rm min}\leq x < \text{HU}_{\rm hard}^{\rm max}\\
    \lambda_{\rm other} |x - \hat{x}| & \text{otherwise}
  \end{cases},
\end{align} where $\text{HU}_{\rm soft}^{\rm min},\text{HU}_{\rm soft}^{\rm max}, \text{HU}_{\rm hard}^{\rm min},\text{ and HU}_{\rm hard}^{\rm max}$ are chosen as the upper and lower boundaries of soft and hard tissue density values. Then $\mathcal{L}_{\rm mw}({\bm x}, \hat{{\bm x}})$ is the average over the image patch.

In all the other levels, we use the GAN loss with regular $\rm L1$ loss without the multi-window low nor the perceptual loss because the the concept of soft and hard tissues and perceptual similarity are void in latent vector space.

During inference, given input image $\mathbf{X}$, the reconstruction of a patch, ${\bm x}$, is obtained with $\mathcal{D}^{(L)}\circ ...\mathcal{D}^{(1)}(\mathcal{E}^{(1)}\circ...\mathcal{E}^{(L)}({\bm x}))$ the reconstruction of the whole image, $\mathbf{\hat{X}}$, is obtained with sliding window inference\cite{monai_consortium_2024_13942962}. The latent vector of a patch is obtained with ${\bm z}=\mathcal{E}^{(1)}\circ...\mathcal{E}^{(L)}({\bm x})$. Similarly, we use sliding window inference to compute the latent vector of the whole image $\mathbf{Z}$. For simplicity, in the following text, we omit the sliding window inference and denote the reconstruction and latent vector of the whole image as $\mathbf{\hat{X}}=\mathcal{D}^{(L)}\circ ...\mathcal{D}^{(1)}(\mathcal{E}^{(1)}\circ...\mathcal{E}^{(L)}(\mathbf{X}))$ and $\mathbf{Z}=\mathcal{E}^{(1)}\circ...\mathcal{E}^{(L)}(\mathbf{ X})$

We use a single-stacked autoencoder as our segmentation autoencoder
$(\mathcal{E}_{\rm seg},\mathcal{D}_{\rm seg})$.
We use the dice coefficient loss $\mathcal{L}_{\rm dice}$ as the reconstruction loss.

\subsubsection{Attribute Conditioned Latent Diffusion Model}
Similar to Patrick et al.\cite{khader2023denoising}, we built upon the 2D U-Net based latent diffusion model developed by Esser et al.\cite{esser2020taming} for natural image generation and developed a 3D U-Net latent diffusion model. The 2D operations in the 2D U-Net were propagated to 3D operations to support 3D latent diffusion.

The classifier free guidance \cite{ho2022classifier} was used for attribute conditioning. In our study, we consider categorical attributes. A patient's attributes are first converted to categories: $\mathbf{a} = (a_{\rm sex}, a_{\rm age}, a_{\rm height}, a_{\rm weight})$ including sex, age, weight, and height (with an additional $a_{\mathtt{none}}$ category for each attribute) as dicussed in section \ref{sec:model-overview} and then mapped to learnable embeddings. The embeddings are then incorporated to each level of the 3D U-Net to guide the denosing process following Patrick et al\cite{esser2020taming}. The $a_{\mathtt{none}}$ is used to represent unavailable attributes or randomly dropped attributes in classifier free guidance.

Let the latent embeddings of a CT image and segmentation be $\mathbf{Z}_{\rm img}=\mathcal{E}_{\rm img}(\mathbf{X})$, and $\mathbf{Z}_{\rm seg}=\mathcal{E}_{\rm seg}(\mathbf{Y})$, where $\mathbf{X}$ is a CT image and $\mathbf{Y}$ is the segmentation of the CT image $\mathbf{X}$. The latent diffusion model $\epsilon_{\theta}$ takes both as the input the learn the joint distribution of CT and segmentation latent embeddings $\mathbf{Z}=[\mathbf{Z}_{\rm img},\mathbf{Z}_{\rm seg}]$ conditioning on the patient metadata $\mathbf{a}$.

\subsubsection{Post-processing Model}
Since the proposed stacked autoencoder preserves most anatomical structures, the reconstructed images tends to be overly smooth compared with the original image causing the sampled images to be also smooth and lack of details. To restore the lost high frequency information, we trained a 3D U-Net to post-process the decoded images.

Let the post-process 3D U-Net be $f(\mathbf{\hat{X}})$, where $\mathbf{\hat{X}}=\mathcal{D}_{\rm img}(\mathcal{E}_{\rm img}(\mathbf{X}))$ is a reconstructed image. We train $f$ with the $\rm L1$ loss and perceptual loss to minimize the distance between $f(\mathbf{\hat{X}})$ and $\mathbf{X}$. The loss function is defined as:
\begin{align*}
  \mathcal{L}_{\rm post}(\mathbf{\hat{X}},\mathbf{X}) = {\rm L1}(\mathbf{\hat{X}},\mathbf{X})+\mathcal{L}_{per}(\mathbf{\hat{X}},\mathbf{X})
\end{align*}
We then process sampled images $\mathbf{\tilde{X}}$ to restore lost details and increase fidelity with the post process model. The post processed sample image is $\mathbf{\tilde{X}}_{\rm post}=f(\mathbf{\tilde{X}})$

The post-processing mode use identical architecture as the latent diffusion model but without any conditioning.
\subsection{Training Details}

Here, we describe the training details for the above model, including the full-body CT dataset used for training and validation. The downstream models consider during the virtual clinical trial are also described below.

\subsubsection{Full-body CT Dataset}

We derive a dataset of 798 full body CTs from the New Mexico Decedent Image Database (NMDID)\cite{edgar2020new}, an open resource maintained by the University of New Mexico that provides a de-identified CT scans of deceased individuals. This database includes CT scans from over 15,000 de-identified individuals, collected between 2010 and 2017. The standard collection protocol includes three scans that together cover the full body:
\begin{inlist}
\item the head, neck and upper extremities (H-N-UXT);
\item the torso (TORSO); and
\item lower extremities (LEXT).
\end{inlist}
We use organ centroids from TotalSegmentator to initialize a rigid intensity-based registration, keeping the majority of the H-N-UXT scan for the overlapping region. This generally includes the arms, which are folded over the chest. These are then resized to a resolution of $1 \times 1$ mm with a slice thickness of $3$mm.
Segmentations of the body, 128 organs, and 3 tissue types are acquired using TotalSegmentator.\cite{wasserthal2023totalsegmentator}
For the segmentation autoencoder, the 128 organs are reduced to 16 by combining related organs. The ``bone'' class refers to non-appendicular bones. Large organs were prioritized over small organs to capture as much anatomical structure while preserving GPU memory. The full list or organ classes are listed in Table~\ref{tab:organ-seg}.

\subsubsection{VQ-VAE Training Details}
Due to the high memory consumption of 3D convolutions, the autoencoders and post-processing model are implemented in a patch-based manner. Let $\mathbf{x}\in \mathbb{R}^{h\times w\times d}$ denotes a $(h,w,d)$ sized patch of an input image $\mathbf{X}\in \mathbb{R}^{H\times W\times D}$ of size $(H,W,D)$. The embedding of a whole CT image and segmentation $\mathbf{Z}=[\mathbf{Z}_{\rm img}, \mathbf{Z}_{\rm seg}]$ is obtained with sliding window approach \cite{monai_consortium_2024_13942962}.

We developed three types of autoencoders, including vanilla CT image autoencoders, a 2 level stacked CT image autoencoders, and vanilla segmentation autoencoders. The compression rate (or composed compression rate for stacked autoencoders) of all autoencoders is kept at 16 along each dimension. The patch sizes of vanilla CT image autoencoders and segmentation autoencoders are (128,128,128). The stacked CT image autoencoders use (128,128,128) patch sizes at the highest level ($\mathcal{E}^{(2)},\mathcal{D}^{(2)}$), and (96,96,96) at the lowest level ($\mathcal{E}^{(1)},\mathcal{D}^{(1)}$). AdamW \cite{loshchilov2018decoupled} is used as the optimizer with a learning rate of $0.0000375$ for all optimizers. The batch size of all autoencoders is 1. The latent dimensions of the U-Net of the vanilla CT image autoencoders, the 2-level stacked image autoencoders, and the vanilla segmentation autoencoders are $(32,64,128,256)$; $(64,128)$ (level 2) and $(128,256)$ (level 1); and $(32,64,128,256)$.
In our experiments we used a VQ-VAE with a level-2 stacking and a composed downscaling factor of 16 based on the hyperparameter search in Table \ref{tab:hyperparam-search}.

\subsubsection{Latent Diffusion Model}

The latent vectors of the image $\mathbf{Z}_{\rm img}$ and the segmentation $\mathbf{Z}_{\rm seg}$ are concatenated together as $\mathbf{Z}\in \mathbb{R}^{2\times 48\times 48\times 48}$. The latent diffusion model is trained to diffusion and reverse diffusion the latent vector using a U-Net. The latent dimensions of the U-Net are $(160,320,720,1280)$. The learning rate is kept as 1 and batch size is 1. The dimension of each patient attribute embedding is 32. During training, the attribute embedding is set to zero with a probability of 0.2 for classifier free guidance. AdamW\cite{loshchilov2018decoupled} with a learning rate of $0.0001$ is used as the optimizer. 4 NVIDIA A10 GPUs are used to train the latent diffusion model, each with 20GB of GPU memory.

\subsubsection{Post-processing Model}

The post-processing model is also developed to process patches. The patch size is $(80,80,24)$. Same U-Net structure as the latent diffusion model is used with the same latent dimensions. Learning rate is set as $0.0001$ and batch size is also 1. We use AdamW\cite{loshchilov2018decoupled} as the optimizer with a learning rate of $0.0001$. One NVIDIA A5000 with 48GB of GPU memory is used for training. PyTorch\cite{paszke2019pytorch} is used as the deep learning framework for all the experiments in this paper.

\label{sec:vct-methods}
\begin{figure}[h]
  \centering
  \includegraphics[width=\linewidth]{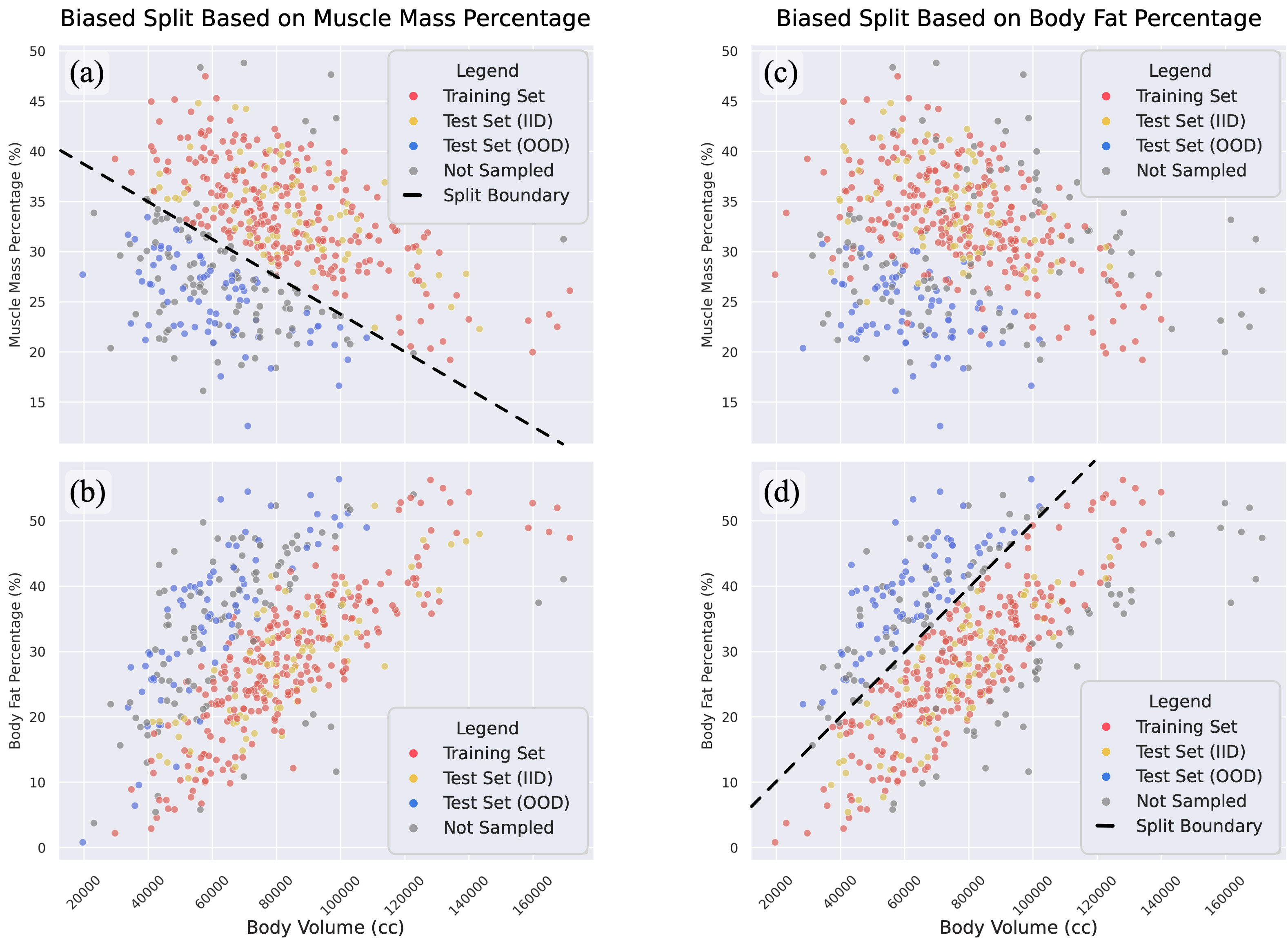}
  \caption{The biased splits used in our virtual clinical trial. To encourage shortcut learning in the downstream model, we choose a linear decision boundary based on body volume. When regressing the body fat percentage, we use the split based on muscle mass percentage (a-b), resulting in biased, but not fully separable, groups for training and ID testing, with simulated deployment.}
  \label{fig:biased-splits}
\end{figure}

\subsection{Downstream AI Model Training}
In this section we describe the data, AI models, and training for the downstream precision medicine tasks discussed in Section~\ref{sec:downstream}. We consider two tasks in body composition measurement, an important part of precision medicine that provides more physically meaningful measurements than the body mass index but may be difficult to measure. The ground truth for all images is obtained through analysis of an independent segmentation, as described in Section~\ref{sec:body-measures}.


\subsubsection{Biased Datasets for Body Composition Measurement}
\label{sec:biased-datasets}

To create a biased model, we intentionally bias the training set based on patient attributes. For body fat percentage regression, we divided the data based on a linear decision boundary in terms of muscle mass and body volume. Because body fat and muscle mass percentage are related, this results in a high correlation between the body volume and body fat percentage (Pearson $r = 0.872$ in the training set). This creates a potential shortcut for model learning to estimate the body fat percentage based on body volume, resulting in a biased output. For muscle mass percentage regression, we take the same approach, while splitting the training distribution and OOD samples based on body fat so that the target variable is not directly used in the split, resulting in a Pearson coefficient of $-0.696$ between muscle mass and body volume. In each case, the training set consists of 200 real CT images, while the ID and OOD test sets contain 75 images each.
We denote each test split by the set of patient identifiers, \eg, $S_{\rm ID}^{\rm (fat)}$ or $S_{\rm OOD}^{\rm fat}$ to denote the ID and OOD test sets of the split based on body fat percentage, respectively.
The downstream model consists of a Swin-B transformer backbone\cite{liu2021swin} with ImageNet-21k pre-training.\cite{ridnik2021imagenet}. The backbone image encoder processes a sagittal and coronal loss, which are then concatenated and followed by a linear layer with scalar output and mean squared error loss. During training, slices are sampled randomly from the middle third of the CT image, while during validation the middle slice is sampled deterministically, and resized to $384 \times 384$. The downstream model is trained for 100 epochs with a batch size of 16 and an initial learning rate of 0.0001, decreased by a factor of 10 at epoch 50 and again at epoch 90, using the AdamW optimizer.

\subsubsection{Importance Weighting Details}

The importance weighting discussed in Section~\ref{sec:vct-degradation} depends on the ability to model the likelihood $p(\text{OOD}|\mathbf{a})$ of a given sample belonging to the OOD population, given the conditioning attributes $\mathbf{a}$.
To implement this approach, we train a random forest classifier to distinguish between ID and OOD samples based on $\mathbf{a}$. The classifier, which has 100 trees and a minimum of 10 samples per leaf, is trained on the ID and OOD test sets. It achieves an accuracy of 0.83 For the BFP regression task and 0.90 for MMP. The likelihood $p(\text{OOD}|\mathbf{a})$ is given by the fraction of trees in the random forest that classify the sample as OOD.
The importance weighted MAE is then computed as
\begin{align}
  \label{eq:importance-weighted-mae}
  \mathcal{L}_{\rm imp} = \frac{1}{\sum_j w_j} \sum_{i} w_i\, |y_i - \hat{y}_i|,
\end{align} where
\begin{align}
  w_i &= \frac{p(\text{OOD}|\mathbf{a}_i)}{p(\text{ID}|\mathbf{a}_i)} \frac{p(\text{ID})}{p(\text{OOD})}\\
  &= \frac{p(\text{OOD}|\mathbf{a}_i)}{1 - p(\text{OOD}|\mathbf{a}_i)} \frac{p(\text{ID})}{p(\text{OOD})}.
\end{align} For $p(\text{ID})$ and $p(\text{OOD})$, we use the proportion of samples from the original image set assigned to each population based on the biased split described in Section~\ref{sec:biased-datasets}.

\backmatter









\section*{Declarations}

\bmhead{Funding}

This work has been supported by Oracle for Research, the Link Foundation for Modeling, Simulation, and Training, Johns Hopkins University internal funds.

\bmhead{Conflict of interest/Competing interests} The authors declare they have no competing interests.

\bmhead{Ethics approval and consent to participate}

No data was collected for this study. The NMDID is an open resource maintained by the University of New Mexico that provides a de-identified CT scans of deceased individuals. The NMDID was reviewed by the University of New Mexico Institutional Review Board (IRB), which determined that IRB approval was not required.

\bmhead{Consent for publication}

\bmhead{Data availability}

The dataset used in this paper is available upon request to and approval by the New Mexico Decedent Image Database \cite{edgar2020new}.

\bmhead{Materials availability}

Not applicable.

\bmhead{Author contribution}
Benjamin D. Killeen and Bohua Wan contributed equally to this work and may list their names first on their respective CVs. Benjamin D. Killeen led the initial conceptualization, generative modeling design, design and implementation of VCTs, manuscript writing, and project management. Bohua Wan led the generative modeling design and implementation, synthetic image validation, and contributed to manuscript writing. Aditya Kulkarni contributed to the design and implementation of generative models and manuscript writing. Nathan Drenkow contributed to the design of VCTs. Michael Oberst contributed to the conceptualization and VCT design. Paul H. Yi provided expertise on clinical background and contributed to VCT design. Mathias Unberath oversaw the conceptualization, project management, and manuscript writing.

\begin{appendices}

  \begin{table}[t]
    \ra{1.3}
    \centering
    \begin{tabular}{lrrrrrrr}
      \toprule
      Class & DICE & \multicolumn{1}{r}{Volume Corr.} & \multicolumn{3}{c}{Centroid Corr.} &  \\
      \cmidrule(lr){4 - 6}
      & && R & A & S \\
      \midrule
      Bone                  & 0.815 $\pm$ 0.023    & 0.996 & 1.000 & 1.000 & 1.000 \\
      Spleen                & 0.661 $\pm$ 0.170    & 0.887 & 0.983 & 0.986 & 1.000 \\
      Kidney                & 0.689 $\pm$ 0.239    & 0.894 & 0.880 & 0.953 & 0.905 \\
      Liver                 & 0.874 $\pm$ 0.089    & 0.985 & 0.999 & 0.998 & 1.000 \\
      Lung upper lobes      & 0.626 $\pm$ 0.135    & 0.835 & 0.784 & 0.989 & 1.000 \\
      Lung lower lobes      & 0.548 $\pm$ 0.122    & 0.831 & 0.757 & 0.985 & 1.000 \\
      Lung middle lobe      & 0.688 $\pm$ 0.117    & 0.856 & 0.998 & 0.992 & 1.000 \\
      Urinary bladder       & 0.675 $\pm$ 0.162    & 0.853 & 0.999 & 0.996 & 1.000 \\
      Prostate              & 0.573 $\pm$ 0.373    & 0.692 & 0.645 & 0.708 & 0.675 \\
      Heart                 & 0.799 $\pm$ 0.103    & 0.963 & 0.993 & 0.998 & 1.000 \\
      Aorta                 & 0.563 $\pm$ 0.137    & 0.818 & 0.998 & 0.990 & 0.999 \\
      Gluteus Muscles       & 0.609 $\pm$ 0.026    & 0.992 & 0.712 & 0.995 & 1.000 \\
      Autochthonous Muscles & 0.893 $\pm$ 0.022    & 0.997 & 1.000 & 1.000 & 1.000 \\
      Iliopsoas             & 0.820 $\pm$ 0.057    & 0.991 & 0.999 & 1.000 & 1.000 \\
      Brain                 & 0.964 $\pm$ 0.009    & 0.998 & 1.000 & 0.999 & 1.000 \\
      Appendicular Bones    & 0.839 $\pm$ 1.048    & 0.993 & 0.999 & 0.995 & 0.975 \\
      \cmidrule(lr){1-1}
      Average & 0.727 $\pm$ 0.115 &  0.911 & 0.922 & 0.974  & 0.972 \\
      \bottomrule
    \end{tabular}
    \caption{Anatomical Consistency of Synthetic Images per Organ}
    \label{tab:organ-seg}
  \end{table}

  \section{Additional Experiments}
  \label{sec:additional-experiments}

  \subsection{Qualitative Results}

  Fig.~\ref{fig:qualitative-sample} shows a real image, corresponding reconstruction, and synthetic sample generated by our model, using the same patient attributes. Qualitatively, the reconstruction closely aligns with the original image, including an amputated right leg. The synthetic image, which is generated from the same patient attributes but otherwise has no additional information about the real image, is sufficiently realistic to be segmented by TotalSegmentator\cite{wasserthal2023totalsegmentator}. There are noted failure modes for complicated or subtle structures, which may be inconsistent in our training data. For example, synthetic images like the one in Fig.~\ref{fig:qualitative-sample} (c-e) often feature distorted arm bones because the scans were acquired without consistently placing the left arm over right, or vice versa. Other areas where our model falls short include the topology of the rib, intestines, and other connected structures. Although these issues may be resolved by augmenting our training data with additional partial or full-body images, the synthetic images produced by our model are of sufficient quality to support VCTs, assuming they produce images that align with the conditioning parameters, so as to produce a virtual patient cohort with the desired attributes.

  \subsection{Hyperparameter Search and Ablation Studies}
  \label{sec:ablation-studies}

  To determine the best configuration for our generative model, we conducted a series of ablation studies and hyperparameter searches. First, Table~\ref{tab:reconstruction-ablation} shows the effect of stacking autoencoder layers and the multi-window L1 loss on reconstruction quality, a necessary capability for the generative model to produce visually realistic images.
  The reconstruction performances are evaluated in terms of Peak signal-to-noise ratio (PSNR) and structural similarity index measure (SSIM). As discussed in Section~\ref{sec:stacked-autoencoder}, when the stack level $L_{ae}$ is 1, the autoencoder is simply the vanilla VQ-VAE, resulting in a PSNR of 29.08 and SSIM of 0.9337. With $L_{ae}=2$, the PSNR and SSIM improve to 30.97 and 0.9494, indicating better reconstruction quality. Using multi-window loss $\mathcal{L}_{mw}$ slightly improves reconstruction quality for vanilla VQ-VAE (with $L_{ae}=1$) as suggested by the higher PSNR and SSIM scores. However, PSNR decreased and SSIM decreased when using $\mathcal{L}_{mw}$ with stacked autoencoders. We hypothesize that the inconsistency between PSNR and SSIM trend is because the multi-window loss is only used in the top layer of the stacked autoencoder, which is a VQ-VAE with 4 times compression and without using multi-window the VQ-VAE is already capable of performing 4 times compression well. Based on the advantage of multi-window loss for vanilla VQ-VAE, we use multi-window loss for the stacked autoencoder in the following experiments.

  We conduct a hyperparameter search over the codebook size of the VQ-VAE, as shown in Table~\ref{tab:hyperparam-search}. The results show that a codebook size of 4096 works best in terms of SSIM with both vanilla VQ-VAE and stacked autoencoder. Using 4096 as the codebook size, vanilla VQ-VAE also achieves the best PSNR. The stacked autoencoder with $L_{ae} = 2$ achieves the best PSNR with codebook size of 2048, however, the difference between codebook size of 2048 and 4096 are small in terms of PSNR scores. We use 4096 as the codebook size in our main model.

  Having determined an appropriate codebook size and loss function, we conducted an ablation study to evaluate the advantage of our model's main components on synthetic image quality.
  As discussed in Section~\ref{sec:realism}, we use the Frechet Inception Distance (FID) to quantify the realism of synthetic images, taking a subset of training images as the real image reference set.
  Because FID is a metric designed for 2D images, we compute FID scores for slices from the full body images as well as cropped portions of the body, including the head and neck region, the torso, and the lower extremities.
  As can be seen in Table~\ref{tab:fid-scores}, considerable improvements are consistently observed when comparing latent diffusion models using stacked autoencoders for reconstruction with those using vanilla VQ-VAE, bringing the average FID from 34.04 to 10.07 without joint segmentation modeling and 10.26 with. This shows that jointly learning the anatomical segmentation, which is necessary to produce images with anatomical realism, has little effect on low-level image realism. Application of a further post-processing network further improves average FID to 5.97.

  \begin{table}[]
    \centering
    \caption{Reconstruction Ablation Study}
    \begin{tabular}{rcrr}
      \toprule
      $L_{\rm ae}$ & $\mathcal{L}_{\rm mw}$ & PSNR & SSIM \\
      \midrule
      1 & \xmark & 29.08 & 0.9337\\
      1 & \cmark & 29.35 & 0.9371\\
      2 & \xmark &  30.97 & \textbf{0.9494}\\
      2 & \cmark & \textbf{31.06} & 0.9472\\
      \bottomrule
    \end{tabular}
    \label{tab:reconstruction-ablation}
  \end{table}

  \begin{table*}[]
    \centering
    \caption{Hyperparameter Search over Codebook Size}
    \begin{tabular}{rrrr}
      \toprule
      $L_{\rm ae}$ & Codebook Size & PSNR & SSIM \\
      \midrule
      1 & 512 & 29.10 & 0.9309\\
      1 & 1024 & 29.48 & 0.9333\\
      1 & 2048 & 29.35 & 0.9371\\
      1 & 4096 & 29.89 & 0.9385\\
      1 & 8192 & 29.08 & 0.9337\\
      2 & 512 & 30.96 & 0.9430\\
      2 & 1024 & 31.05 & 0.9442\\
      2 & 2048 & \textbf{31.11} & 0.9467\\
      2 & 4096 & 31.06 & \textbf{0.9472}\\
      2 & 8192 & 30.96 & 0.9458\\
      \bottomrule
    \end{tabular}
    \label{tab:hyperparam-search}
  \end{table*}

  \begin{sidewaystable}[]
    \ra{1.3}
    \centering
    \caption{}
    \begin{tabular}{rrcrrrrrrrrrrrrrrrr}
      \toprule
      &&& \multicolumn{4}{c}{Head \& Neck (FID)} & \multicolumn{4}{c}{Torso (FID)} & \multicolumn{4}{c}{Lower Extremities (FID)} & \multicolumn{4}{c}{Full Body (FID)} \\
      \cmidrule(lr){4-7} \cmidrule(lr){8-11} \cmidrule(lr){12-15} \cmidrule(lr){16-19}
      \multicolumn{1}{c}{$L_{\rm AE}$}& \multicolumn{1}{c}{$\mathcal{L}_{\rm post}$} & Seg. & Sag. & Cor. & Ax. & Avg. & Sag. & Cor. & Ax.  & Avg. & Sag. & Cor. & Ax.  & Avg. & Sag. & Cor. & Ax.  & Avg. \\
      \midrule
      1 & 1 & \cmark & 26.80 & 19.01 & 28.05 & 24.62 & 49.19 & 34.91 & 31.52 & 38.54 & 37.18 & 28.29 & 34.81 & 33.43 & 34.85 & 25.76 & 41.50 & 34.04\\
      2 & 1 & \xmark & 6.70 & 7.31 & 2.69 & 5.57 & 17.86 & 12.88 & 8.44 & 13.06 & 16.09 & 13.58 & 14.17 & 14.61 & 10.15 & 6.94 & 13.13 & 10.07\\
      2 & 1 & \cmark & 8.14 & 8.62 & 3.49 & 6.75 & 19.26 & 13.34 & 11.08 & 14.56 & 14.21 & 10.97 & 14.31 & 13.16 & 10.26 & 6.88 & 13.64 & 10.26\\
      2 & 2 & \cmark & 7.95 & 5.39 & 8.30 & 7.21 & 15.89 & 9.40 & 8.63 & 11.31 & 7.98 & 5.39 & 8.30 & 7.22 & 5.84 & 4.22 & 7.85 & 5.97\\
      \bottomrule
    \end{tabular}
    \label{tab:fid-scores}
  \end{sidewaystable}

  \section{Body Measurement Details}
  \label{sec:body-measures}

  \subsubsection{Measuring Mass}
  \label{sec:hu-conversion}

  Density ($\rho$) can be represented using Hounsfield Units (HU) and the mass-normalized HU variant, which is a material's fundamental property.\cite{sudhyadhom2020on}
  \begin{equation}
    \label{eq:B1}
    \text{$\rho$} = (HU + 1000) / (HU_{\rho} + 1000)
  \end{equation}
  Optionally, we can reduce noise by smoothing the air density with typical air HU values to improve the accuracy of density calculations in air-filled regions.
  \[
    \text{HU}_{\text{adjusted}} =
    \begin{cases}
      -1000, & \text{if } \text{HU} \leq -900 \\
      \text{HU}, & \text{otherwise}
    \end{cases}
  \]
  Tissue-specific densities are then calculated for each tissue type for various segmented regions (such as adipose, muscle, liver, and bone) using equation \ref{eq:B1}. Density values across all body regions are summed after applying a body segmentation mask to calculate body mass. This total density is then multiplied by the voxel volume to obtain body mass in grams, serving as a basis for further metrics such as fat, muscle, and bone mass for comprehensive body composition analysis. The total density is summed over the full body segmentation to obtain the body weight in kilograms. As shown in Fig.~\ref{fig:calibration-weight}, the body weight measured in this manner is highly correlated with the body weight recorded at the time of measurement, with an $R^2$ value of 0.95. The systematic error between the two may arise because of different amounts of clothing worn at the time of imaging, which are not included in the body segmentation.

  \subsubsection{Measuring Height}
  \label{sec:height}
  \begin{figure*}[t]
    \centering
    \centering
    \subfloat[Height Calibration]{
      \includegraphics[width=0.5\textwidth]{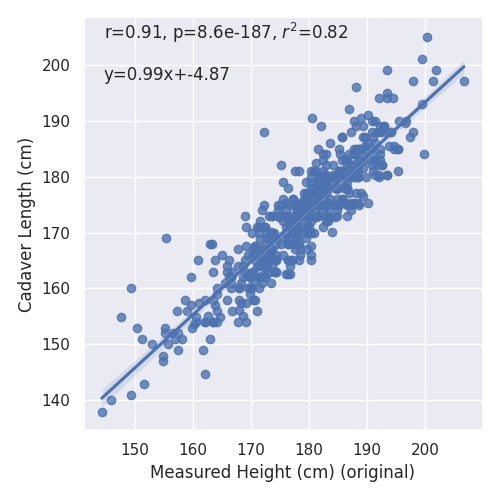}
      \label{fig:calibration-height}
    }
    \subfloat[Weight Calibration]{
      \includegraphics[width=0.5\textwidth]{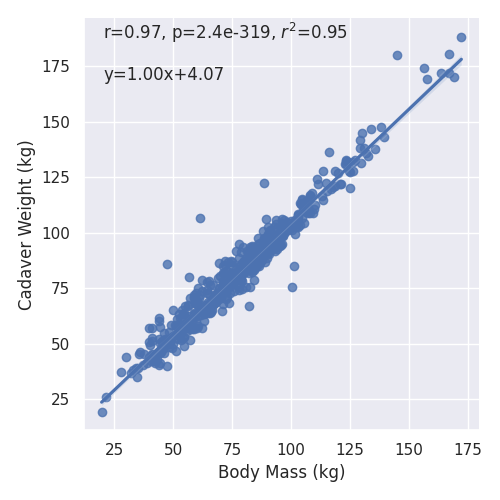}
      \label{fig:calibration-weight}
    }
    \caption{The difference between computed height (a) and weight (b) using the automatic pipelines here and the recorded height and weight of cadaveric specimens at the time of imaging. We correct for the systematic error between these two measurement types in our validation experiments.}
    \label{fig:calibration}
  \end{figure*}

  Body height can be difficult to measure consistently. For living individuals, it can be affected by the posture, time of day, and method of measurement.\cite{wang2012measurement} For cadaveric specimens, such as those used in this study, measuring height from full-body images in a manner that aligns with living height is nontrivial, due to specimens' variable orientation and pose. Even with reliable full body segmentation, simply measuring the distance from feet to head is not sufficient, because the feet are generally pointed due to gravity, rather than angled as they would be when standing. These properties are likewise reflected in synthetic images, making it necessary to compute height by dividing the body into pose-based segments based on multi-organ segmentation~\cite{wasserthal2023totalsegmentator}.
  We first extract watertight meshes for each segmentation mask using marching cubes and orient the anatomy with the true RAS coordinate system of the anatomy, taking the superior axis from principle component analysis (PCA) of the full body mesh vertices. The left/right axis is approximated from the average difference between symmetrical organs, such as the halves of the pelvis, the clavicles, and the scapula, and the basis is completed according to Gram-Schmidt orthogonalization.
  The lower segment of the body is defined by a ``pelvis plane,'' normal to the superior axis and at the superior-most point of the femurs. Because each leg may be bent differently, the leg heights are computed independently for the left and right sides, and the maximum is used.
  The inferior-most point of the femur is identified, and tibia parts are split, with components above the femur’s inferior point excluded. A knee plane is defined at the superior-most point of the tibia. PCA on the tibia mesh vertices yields its long axis. The intersection of this axis with the body mesh, on the underside of the foot, gives the length of the lower leg. The upper leg length is given by the long axis of the femur, between the knee plane and the pelvis plane. The total leg height is then the sum of the lower and upper leg lengths, and the lower body segment length is the maximum of the two leg lengths.
  Torso height is determined by measuring the distance along the superior axis between the pelvis plane and the centroid of the C7 vertebra.
  The neck height is computed as the length of the segment between the centroids of the C7 and C1 vertebrae. For the head height, a line is projected from the centroid of the C1 vertebra towards the C2 vertebra, and the segment extending from the C1 vertebra to the crown of the head is used.
  The total height is computed as the sum of the lower body, torso, neck, and head lengths.
  Fig.~\ref{fig:calibration-height} shows the difference between heights computed in this manner and the original cadaver height in the NMDID training set, as measured postmortem. In our experiments, we correct for the difference between the based on a linear fit.




\end{appendices}

\bibliography{ref}

\begin{thebibliography}{10}
\expandafter\ifx\csname url\endcsname\relax
  \def\url#1{\burl{#1}}\fi
\expandafter\ifx\csname urlprefix\endcsname\relax\def\urlprefix{URL }\fi
\providecommand{\bibinfo}[2]{#2}
\providecommand{\eprint}[2][]{\url{#2}}
\providecommand{\doi}[1]{\url{https://doi.org/#1}}
\bibcommenthead

\bibitem{titano2018automated}
\bibinfo{author}{Titano, J.~J.} \emph{et~al.}
\newblock \bibinfo{title}{Automated deep-neural-network surveillance of cranial images for acute neurologic events}.
\newblock \emph{\bibinfo{journal}{Nature medicine}} \textbf{\bibinfo{volume}{24}}, \bibinfo{pages}{1337--1341} (\bibinfo{year}{2018}).

\bibitem{levin2018machine}
\bibinfo{author}{Levin, S.} \emph{et~al.}
\newblock \bibinfo{title}{Machine-learning-based electronic triage more accurately differentiates patients with respect to clinical outcomes compared with the emergency severity index}.
\newblock \emph{\bibinfo{journal}{Annals of emergency medicine}} \textbf{\bibinfo{volume}{71}}, \bibinfo{pages}{565--574} (\bibinfo{year}{2018}).

\bibitem{liu2019comparison}
\bibinfo{author}{Liu, X.} \emph{et~al.}
\newblock \bibinfo{title}{A comparison of deep learning performance against health-care professionals in detecting diseases from medical imaging: a systematic review and meta-analysis}.
\newblock \emph{\bibinfo{journal}{The lancet digital health}} \textbf{\bibinfo{volume}{1}}, \bibinfo{pages}{e271--e297} (\bibinfo{year}{2019}).

\bibitem{pickhardt2013opportunistic}
\bibinfo{author}{Pickhardt, P.~J.} \emph{et~al.}
\newblock \bibinfo{title}{Opportunistic screening for osteoporosis using abdominal computed tomography scans obtained for other indications}.
\newblock \emph{\bibinfo{journal}{Annals of internal medicine}} \textbf{\bibinfo{volume}{158}}, \bibinfo{pages}{588--595} (\bibinfo{year}{2013}).

\bibitem{jang2019opportunistic}
\bibinfo{author}{Jang, S.} \emph{et~al.}
\newblock \bibinfo{title}{Opportunistic osteoporosis screening at routine abdominal and thoracic ct: normative l1 trabecular attenuation values in more than 20 000 adults}.
\newblock \emph{\bibinfo{journal}{Radiology}} \textbf{\bibinfo{volume}{291}}, \bibinfo{pages}{360--367} (\bibinfo{year}{2019}).

\bibitem{eng2021automated}
\bibinfo{author}{Eng, D.} \emph{et~al.}
\newblock \bibinfo{title}{Automated coronary calcium scoring using deep learning with multicenter external validation}.
\newblock \emph{\bibinfo{journal}{NPJ digital medicine}} \textbf{\bibinfo{volume}{4}}, \bibinfo{pages}{88} (\bibinfo{year}{2021}).

\bibitem{oh2024evaluation}
\bibinfo{author}{Oh, S.} \emph{et~al.}
\newblock \bibinfo{title}{{Evaluation of deep learning-based quantitative computed tomography for opportunistic osteoporosis screening}}.
\newblock \emph{\bibinfo{journal}{Sci. Rep.}} \textbf{\bibinfo{volume}{14}}, \bibinfo{pages}{1--9} (\bibinfo{year}{2024}).

\bibitem{zopfs2020evaluating}
\bibinfo{author}{Zopfs, D.} \emph{et~al.}
\newblock \bibinfo{title}{{Evaluating body composition by combining quantitative spectral detector computed tomography and deep learning-based image segmentation}}.
\newblock \emph{\bibinfo{journal}{Eur. J. Radiol.}} \textbf{\bibinfo{volume}{130}}, \bibinfo{pages}{109153} (\bibinfo{year}{2020}).

\bibitem{ozsahin2020review}
\bibinfo{author}{Ozsahin, I.}, \bibinfo{author}{Sekeroglu, B.}, \bibinfo{author}{Musa, M.~S.}, \bibinfo{author}{Mustapha, M.~T.} \& \bibinfo{author}{Ozsahin, D.~U.}
\newblock \bibinfo{title}{{Review on Diagnosis of COVID-19 from Chest CT Images Using Artificial Intelligence}}.
\newblock \emph{\bibinfo{journal}{Comput. Math. Methods Med.}} \textbf{\bibinfo{volume}{2020}}, \bibinfo{pages}{9756518} (\bibinfo{year}{2020}).

\bibitem{murugesan2022a}
\bibinfo{author}{Murugesan, M.} \emph{et~al.}
\newblock \bibinfo{title}{{A Hybrid deep learning model for effective segmentation and classification of lung nodules from CT images}}.
\newblock \emph{\bibinfo{journal}{J. Intell. Fuzzy Syst.}} \textbf{\bibinfo{volume}{42}}, \bibinfo{pages}{2667--2679} (\bibinfo{year}{2022}).

\bibitem{mohammadi2024deep}
\bibinfo{author}{Mohammadi, S.} \emph{et~al.}
\newblock \bibinfo{title}{{Deep learning-based detection of coronary artery calcification in non-contrast and contrast-enhanced CT scans}} (\bibinfo{year}{2024}).
\newblock \bibinfo{note}{[Online; accessed 18. Jun. 2024]}.

\bibitem{benjamens2020state}
\bibinfo{author}{Benjamens, S.}, \bibinfo{author}{Dhunnoo, P.} \& \bibinfo{author}{Mesk{\'o}, B.}
\newblock \bibinfo{title}{The state of artificial intelligence-based fda-approved medical devices and algorithms: an online database}.
\newblock \emph{\bibinfo{journal}{NPJ digital medicine}} \textbf{\bibinfo{volume}{3}}, \bibinfo{pages}{118} (\bibinfo{year}{2020}).

\bibitem{fda2024artificial}
\bibinfo{title}{{Artificial Intelligence and Machine Learning (AI/ML)-Enabled Medical D}} (\bibinfo{year}{2024}).
\newblock \urlprefix\url{https://www.fda.gov/medical-devices/software-medical-device-samd/artificial-intelligence-and-machine-learning-aiml-enabled-medical-devices}.
\newblock \bibinfo{note}{[Online; accessed 27. Jan. 2025]}.

\bibitem{drenkow2021systematic}
\bibinfo{author}{Drenkow, N.}, \bibinfo{author}{Sani, N.}, \bibinfo{author}{Shpitser, I.} \& \bibinfo{author}{Unberath, M.}
\newblock \bibinfo{title}{A systematic review of robustness in deep learning for computer vision: Mind the gap?}
\newblock \emph{\bibinfo{journal}{arXiv preprint arXiv:2112.00639}}  (\bibinfo{year}{2021}).

\bibitem{Ong_Ly2024-jh}
\bibinfo{author}{Ong~Ly, C.} \emph{et~al.}
\newblock \bibinfo{title}{Shortcut learning in medical {AI} hinders generalization: method for estimating {AI} model generalization without external data}.
\newblock \emph{\bibinfo{journal}{npj Digital Medicine}} \textbf{\bibinfo{volume}{7}}, \bibinfo{pages}{1--10} (\bibinfo{year}{2024}).

\bibitem{Drenkow2023-qj}
\bibinfo{author}{Drenkow, N.~G.} \& \bibinfo{author}{Unberath, M.}
\newblock \bibinfo{title}{{RobustCLEVR}: A benchmark and framework for evaluating robustness in object-centric learning}.
\newblock \emph{\bibinfo{journal}{Proc. IEEE Workshop Appl. Comput. Vis.}} \bibinfo{pages}{4506--4515} (\bibinfo{year}{2023}).

\bibitem{rodrigues2022automated}
\bibinfo{author}{Rodrigues, G.} \emph{et~al.}
\newblock \bibinfo{title}{Automated large artery occlusion detection in stroke: a single-center validation study of an artificial intelligence algorithm}.
\newblock \emph{\bibinfo{journal}{Cerebrovascular Diseases}} \textbf{\bibinfo{volume}{51}}, \bibinfo{pages}{259--264} (\bibinfo{year}{2022}).

\bibitem{rava2021validation}
\bibinfo{author}{Rava, R.~A.} \emph{et~al.}
\newblock \bibinfo{title}{Validation of an artificial intelligence-driven large vessel occlusion detection algorithm for acute ischemic stroke patients}.
\newblock \emph{\bibinfo{journal}{The Neuroradiology Journal}} \textbf{\bibinfo{volume}{34}}, \bibinfo{pages}{408--417} (\bibinfo{year}{2021}).

\bibitem{matsoukas2023ai}
\bibinfo{author}{Matsoukas, S.} \emph{et~al.}
\newblock \bibinfo{title}{Ai software detection of large vessel occlusion stroke on ct angiography: a real-world prospective diagnostic test accuracy study}.
\newblock \emph{\bibinfo{journal}{Journal of Neurointerventional Surgery}} \textbf{\bibinfo{volume}{15}}, \bibinfo{pages}{52--56} (\bibinfo{year}{2023}).

\bibitem{wong2021external}
\bibinfo{author}{Wong, A.} \emph{et~al.}
\newblock \bibinfo{title}{External validation of a widely implemented proprietary sepsis prediction model in hospitalized patients}.
\newblock \emph{\bibinfo{journal}{JAMA internal medicine}} \textbf{\bibinfo{volume}{181}}, \bibinfo{pages}{1065--1070} (\bibinfo{year}{2021}).

\bibitem{voter2021diagnostic}
\bibinfo{author}{Voter, A.~F.}, \bibinfo{author}{Meram, E.}, \bibinfo{author}{Garrett, J.~W.} \& \bibinfo{author}{John-Paul, J.~Y.}
\newblock \bibinfo{title}{Diagnostic accuracy and failure mode analysis of a deep learning algorithm for the detection of intracranial hemorrhage}.
\newblock \emph{\bibinfo{journal}{Journal of the American College of Radiology}} \textbf{\bibinfo{volume}{18}}, \bibinfo{pages}{1143--1152} (\bibinfo{year}{2021}).

\bibitem{small2021ct}
\bibinfo{author}{Small, J.}, \bibinfo{author}{Osler, P.}, \bibinfo{author}{Paul, A.} \& \bibinfo{author}{Kunst, M.}
\newblock \bibinfo{title}{Ct cervical spine fracture detection using a convolutional neural network}.
\newblock \emph{\bibinfo{journal}{American Journal of Neuroradiology}} \textbf{\bibinfo{volume}{42}}, \bibinfo{pages}{1341--1347} (\bibinfo{year}{2021}).

\bibitem{kunst2023real}
\bibinfo{author}{Kunst, M.} \emph{et~al.}
\newblock \bibinfo{title}{Real-world performance of large vessel occlusion cadt ai algorithms-what the stroke team needs to know.}
\newblock \emph{\bibinfo{journal}{Journal of the American College of Radiology: JACR}} \bibinfo{pages}{S1546--1440} (\bibinfo{year}{2023}).

\bibitem{wu2021how}
\bibinfo{author}{Wu, E.} \emph{et~al.}
\newblock \bibinfo{title}{{How medical AI devices are evaluated: limitations and recommendations from an analysis of FDA approvals}}.
\newblock \emph{\bibinfo{journal}{Nat. Med.}} \textbf{\bibinfo{volume}{27}}, \bibinfo{pages}{582--584} (\bibinfo{year}{2021}).

\bibitem{lam2022randomized}
\bibinfo{author}{Lam, T. Y.~T.} \emph{et~al.}
\newblock \bibinfo{title}{{Randomized Controlled Trials of Artificial Intelligence in Clinical Practice: Systematic Review}}.
\newblock \emph{\bibinfo{journal}{J. Med. Internet Res.}} \textbf{\bibinfo{volume}{24}}, \bibinfo{pages}{e37188} (\bibinfo{year}{2022}).

\bibitem{oakden2020hidden}
\bibinfo{author}{Oakden-Rayner, L.}, \bibinfo{author}{Dunnmon, J.}, \bibinfo{author}{Carneiro, G.} \& \bibinfo{author}{Re, C.}
\newblock \bibinfo{title}{{Hidden stratification causes clinically meaningful failures in machine learning for medical imaging}} (\bibinfo{year}{2020}).

\bibitem{arpa2024}
\bibinfo{title}{{ARPA-H launches program to help AI-enabled medical tools maintain peak performance}} (\bibinfo{year}{2024}).
\newblock \urlprefix\url{https://arpa-h.gov/news-and-events/arpa-h-launches-program-help-ai-enabled-medical-tools-maintain-peak-performance}.
\newblock \bibinfo{note}{[Online; accessed 3. Dec. 2024]}.

\bibitem{rajpurkar2022ai}
\bibinfo{author}{Rajpurkar, P.}, \bibinfo{author}{Chen, E.}, \bibinfo{author}{Banerjee, O.} \& \bibinfo{author}{Topol, E.~J.}
\newblock \bibinfo{title}{{AI in health and medicine}}.
\newblock \emph{\bibinfo{journal}{Nat. Med.}} \textbf{\bibinfo{volume}{28}}, \bibinfo{pages}{31--38} (\bibinfo{year}{2022}).

\bibitem{segars2008realistic}
\bibinfo{author}{Segars, W.~P.}, \bibinfo{author}{Mahesh, M.}, \bibinfo{author}{Beck, T.~J.}, \bibinfo{author}{Frey, E.~C.} \& \bibinfo{author}{Tsui, B. M.~W.}
\newblock \bibinfo{title}{{Realistic CT simulation using the 4D XCAT phantom}}.
\newblock \emph{\bibinfo{journal}{Med. Phys.}} \textbf{\bibinfo{volume}{35}}, \bibinfo{pages}{3800--3808} (\bibinfo{year}{2008}).

\bibitem{abadi2020virtual}
\bibinfo{author}{Abadi, E.} \emph{et~al.}
\newblock \bibinfo{title}{{Virtual clinical trials in medical imaging: a review}}.
\newblock \emph{\bibinfo{journal}{J. Med. Imaging}} \textbf{\bibinfo{volume}{7}} (\bibinfo{year}{2020}).

\bibitem{badano2023stochastic}
\bibinfo{author}{Badano, A.} \emph{et~al.}
\newblock \bibinfo{title}{{The stochastic digital human is now enrolling for in silico imaging trials{\ifmmode---\else\textemdash\fi}methods and tools for generating digital cohorts}}.
\newblock \emph{\bibinfo{journal}{Prog. Biomed. Eng.}} \textbf{\bibinfo{volume}{5}}, \bibinfo{pages}{042002} (\bibinfo{year}{2023}).

\bibitem{ibrahim2024generative}
\bibinfo{author}{Ibrahim, M.} \emph{et~al.}
\newblock \bibinfo{title}{{Generative AI for Synthetic Data Across Multiple Medical Modalities: A Systematic Review of Recent Developments and Challenges}}.
\newblock \emph{\bibinfo{journal}{arXiv}}  (\bibinfo{year}{2024}).

\bibitem{hung2023med}
\bibinfo{author}{Hung, A. L.~Y.} \emph{et~al.}
\newblock \bibinfo{title}{{Med-cDiff: Conditional Medical Image Generation with Diffusion Models}}.
\newblock \emph{\bibinfo{journal}{Bioengineering}} \textbf{\bibinfo{volume}{10}}, \bibinfo{pages}{1258} (\bibinfo{year}{2023}).

\bibitem{khader2023denoising}
\bibinfo{author}{Khader, F.} \emph{et~al.}
\newblock \bibinfo{title}{{Denoising diffusion probabilistic models for 3D medical image generation}}.
\newblock \emph{\bibinfo{journal}{Sci. Rep.}} \textbf{\bibinfo{volume}{13}}, \bibinfo{pages}{1--12} (\bibinfo{year}{2023}).

\bibitem{chen2024medical}
\bibinfo{author}{Chen, W.} \emph{et~al.}
\newblock \bibinfo{title}{{Medical Image Synthesis via Fine-Grained Image-Text Alignment and Anatomy-Pathology Prompting}} (\bibinfo{year}{2024}).

\bibitem{gao2023synthex}
\bibinfo{author}{Gao, C.} \emph{et~al.}
\newblock \bibinfo{title}{Synthetic data accelerates the development of generalizable learning-based algorithms for x-ray image analysis}.
\newblock \emph{\bibinfo{journal}{Nature Machine Intelligence}}  (\bibinfo{year}{2023}).

\bibitem{hennessy2015the}
\bibinfo{author}{Hennessy, D.~A.} \emph{et~al.}
\newblock \bibinfo{title}{{The Population Health Model (POHEM): an overview of rationale, methods and applications}}.
\newblock \emph{\bibinfo{journal}{Popul. Health Metrics}} \textbf{\bibinfo{volume}{13}}, \bibinfo{pages}{1--12} (\bibinfo{year}{2015}).

\bibitem{guo2024maisi}
\bibinfo{author}{Guo, P.} \emph{et~al.}
\newblock \bibinfo{title}{{MAISI: Medical AI for Synthetic Imaging}}.
\newblock \emph{\bibinfo{journal}{arXiv}}  (\bibinfo{year}{2024}).

\bibitem{esser2020taming}
\bibinfo{author}{Esser, P.}, \bibinfo{author}{Rombach, R.} \& \bibinfo{author}{Ommer, B.}
\newblock \bibinfo{title}{Taming transformers for high-resolution image synthesis} (\bibinfo{year}{2020}).
\newblock \eprint{2012.09841}.

\bibitem{rombach2022high}
\bibinfo{author}{Rombach, R.}, \bibinfo{author}{Blattmann, A.}, \bibinfo{author}{Lorenz, D.}, \bibinfo{author}{Esser, P.} \& \bibinfo{author}{Ommer, B.}
\newblock \bibinfo{title}{{High-Resolution Image Synthesis With Latent Diffusion Models}} (\bibinfo{year}{2022}).
\newblock \urlprefix\url{https://openaccess.thecvf.com/content/CVPR2022/html/Rombach_High-Resolution_Image_Synthesis_With_Latent_Diffusion_Models_CVPR_2022_paper.html}.
\newblock \bibinfo{note}{[Online; accessed 22. Jul. 2024]}.

\bibitem{wasserthal2023totalsegmentator}
\bibinfo{author}{Wasserthal, J.} \emph{et~al.}
\newblock \bibinfo{title}{{TotalSegmentator: Robust Segmentation of 104 Anatomic Structures in CT Images}}.
\newblock \emph{\bibinfo{journal}{Radiology: Artificial Intelligence}}  (\bibinfo{year}{2023}).
\newblock \urlprefix\url{https://pubs.rsna.org/doi/10.1148/ryai.230024}.

\bibitem{heusel2017gans}
\bibinfo{author}{Heusel, M.}, \bibinfo{author}{Ramsauer, H.}, \bibinfo{author}{Unterthiner, T.}, \bibinfo{author}{Nessler, B.} \& \bibinfo{author}{Hochreiter, S.}
\newblock \bibinfo{title}{{GANs Trained by a Two Time-Scale Update Rule Converge to a Local Nash Equilibrium}}.
\newblock \emph{\bibinfo{journal}{Advances in Neural Information Processing Systems}} \textbf{\bibinfo{volume}{30}} (\bibinfo{year}{2017}).
\newblock \urlprefix\url{https://papers.nips.cc/paper/2017/hash/8a1d694707eb0fefe65871369074926d-Abstract.html}.

\bibitem{asadi2021artificial}
\bibinfo{author}{Asadi, F.} \& \bibinfo{author}{O’Reilly, J.~A.}
\newblock \bibinfo{title}{Artificial computed tomography images with progressively growing generative adversarial network} \bibinfo{pages}{1--5} (\bibinfo{year}{2021}).

\bibitem{oreilly2021pretrained}
\bibinfo{author}{O'Reilly, J.~A.} \& \bibinfo{author}{Asadi, F.}
\newblock \bibinfo{title}{Pre-trained vs. random weights for calculating fréchet inception distance in medical imaging} \bibinfo{pages}{1--4} (\bibinfo{year}{2021}).

\bibitem{gatidis2022autopet}
\bibinfo{author}{Gatidis, S.} \emph{et~al.}
\newblock \bibinfo{title}{{A whole-body FDG-PET/CT Dataset with manually annotated Tumor Lesions}}.
\newblock \emph{\bibinfo{journal}{Sci. Data}} \textbf{\bibinfo{volume}{9}}, \bibinfo{pages}{1--7} (\bibinfo{year}{2022}).

\bibitem{leichter1981effect}
\bibinfo{author}{Leichter, I.} \emph{et~al.}
\newblock \bibinfo{title}{The effect of age and sex on bone density, bone mineral content and cortical index}.
\newblock \emph{\bibinfo{journal}{Clinical Orthopaedics and Related Research{\textregistered}}} \textbf{\bibinfo{volume}{156}}, \bibinfo{pages}{232--239} (\bibinfo{year}{1981}).

\bibitem{bates2022ct}
\bibinfo{author}{Bates, D. D.~B.}, \bibinfo{author}{Pickhardt, P.~J.}, \bibinfo{author}{Bates, D. D.~B.} \& \bibinfo{author}{Pickhardt, P.~J.}
\newblock \bibinfo{title}{{CT-Derived Body Composition Assessment as a Prognostic Tool in Oncologic Patients: From Opportunistic Research to Artificial Intelligence{\textendash}Based Clinical Implementation}}.
\newblock \emph{\bibinfo{journal}{Am. J. Roentgenol.}}  (\bibinfo{year}{2022}).

\bibitem{isensee2021nnu}
\bibinfo{author}{Isensee, F.}, \bibinfo{author}{Jaeger, P.~F.}, \bibinfo{author}{Kohl, S. A.~A.}, \bibinfo{author}{Petersen, J.} \& \bibinfo{author}{Maier-Hein, K.~H.}
\newblock \bibinfo{title}{{nnU-Net: a self-configuring method for deep learning-based biomedical image segmentation}}.
\newblock \emph{\bibinfo{journal}{Nat. Methods}} \textbf{\bibinfo{volume}{18}}, \bibinfo{pages}{203--211} (\bibinfo{year}{2021}).

\bibitem{avesta2023comparing}
\bibinfo{author}{Avesta, A.} \emph{et~al.}
\newblock \bibinfo{title}{{Comparing 3D, 2.5D, and 2D Approaches to Brain Image Auto-Segmentation}}.
\newblock \emph{\bibinfo{journal}{Bioengineering}} \textbf{\bibinfo{volume}{10}}, \bibinfo{pages}{181} (\bibinfo{year}{2023}).

\bibitem{stevens2016prediction}
\bibinfo{author}{Stevens, J.}, \bibinfo{author}{Ou, F.-S.}, \bibinfo{author}{Cai, J.}, \bibinfo{author}{Heymsfield, S.~B.} \& \bibinfo{author}{Truesdale, K.~P.}
\newblock \bibinfo{title}{{Prediction of percent body fat measurements in Americans 8 years and older}}.
\newblock \emph{\bibinfo{journal}{Int. J. Obes.}} \textbf{\bibinfo{volume}{40}}, \bibinfo{pages}{587--594} (\bibinfo{year}{2016}).

\bibitem{rathnayake2020development}
\bibinfo{author}{Rathnayake, N.}, \bibinfo{author}{Alwis, G.}, \bibinfo{author}{Lenora, J.} \& \bibinfo{author}{Lekamwasam, S.}
\newblock \bibinfo{title}{{Development and Cross-Validation of Anthropometric Predictive Equations to Estimate Total Body Fat Percentage in Adult Women in Sri Lanka}}.
\newblock \emph{\bibinfo{journal}{Journal of Obesity}} \textbf{\bibinfo{volume}{2020}}, \bibinfo{pages}{2087346} (\bibinfo{year}{2020}).

\bibitem{talma2013bioelectrical}
\bibinfo{author}{Talma, H.} \emph{et~al.}
\newblock \bibinfo{title}{{Bioelectrical impedance analysis to estimate body composition in children and adolescents: a systematic review and evidence appraisal of validity, responsiveness, reliability and measurement error}}.
\newblock \emph{\bibinfo{journal}{Obes. Rev.}} \textbf{\bibinfo{volume}{14}}, \bibinfo{pages}{895--905} (\bibinfo{year}{2013}).

\bibitem{cdc2009quickstats}
\bibinfo{title}{{QuickStats: Mean Percentage Body Fat,{$\ast$} by Age Group and Sex --- National Health and Nutrition Examination Survey, United States, 1999--2004}} (\bibinfo{year}{2009}).
\newblock \urlprefix\url{https://www.cdc.gov/mmwr/preview/mmwrhtml/mm5751a4.htm}.
\newblock \bibinfo{note}{[Online; accessed 4. Feb. 2025]}.

\bibitem{janssen2000skeletal}
\bibinfo{author}{Janssen, I.}, \bibinfo{author}{Heymsfield, S.~B.}, \bibinfo{author}{Wang, Z.} \& \bibinfo{author}{Ross, R.}
\newblock \bibinfo{title}{{Skeletal muscle mass and distribution in 468 men and women aged 18{\textendash}88 yr}}.
\newblock \emph{\bibinfo{journal}{J. Appl. Physiol.}}  (\bibinfo{year}{2000}).

\bibitem{pavlak2023data}
\bibinfo{author}{Pavlak, M.}, \bibinfo{author}{Drenkow, N.}, \bibinfo{author}{Petrick, N.}, \bibinfo{author}{Farhangi, M.~M.} \& \bibinfo{author}{Unberath, M.}
\newblock \bibinfo{title}{{Data AUDIT: Identifying Attribute Utility- and Detectability-Induced Bias in Task Models}} (\bibinfo{year}{2023}).

\bibitem{cho2024medisyn}
\bibinfo{author}{Cho, J.} \emph{et~al.}
\newblock \bibinfo{title}{{MediSyn: Text-Guided Diffusion Models for Broad Medical 2D and 3D Image Synthesis}}.
\newblock \emph{\bibinfo{journal}{arXiv}}  (\bibinfo{year}{2024}).

\bibitem{van2017neural}
\bibinfo{author}{Van Den~Oord, A.}, \bibinfo{author}{Vinyals, O.} \emph{et~al.}
\newblock \bibinfo{title}{Neural discrete representation learning}.
\newblock \emph{\bibinfo{journal}{Advances in neural information processing systems}} \textbf{\bibinfo{volume}{30}} (\bibinfo{year}{2017}).

\bibitem{monai_consortium_2024_13942962}
\bibinfo{author}{Consortium, M.}
\newblock \bibinfo{title}{Monai: Medical open network for ai} (\bibinfo{year}{2024}).
\newblock \urlprefix\url{https://doi.org/10.5281/zenodo.13942962}.

\bibitem{johnson2016perceptual}
\bibinfo{author}{Johnson, J.}, \bibinfo{author}{Alahi, A.} \& \bibinfo{author}{Fei-Fei, L.}
\newblock \bibinfo{editor}{Leibe, B.}, \bibinfo{editor}{Matas, J.}, \bibinfo{editor}{Sebe, N.} \& \bibinfo{editor}{Welling, M.} (eds) \emph{\bibinfo{title}{Perceptual losses for real-time style transfer and super-resolution}}.
\newblock (eds \bibinfo{editor}{Leibe, B.}, \bibinfo{editor}{Matas, J.}, \bibinfo{editor}{Sebe, N.} \& \bibinfo{editor}{Welling, M.}) \emph{\bibinfo{booktitle}{Computer Vision -- ECCV 2016}}, \bibinfo{pages}{694--711} (\bibinfo{publisher}{Springer International Publishing}, \bibinfo{address}{Cham}, \bibinfo{year}{2016}).

\bibitem{ho2022classifier}
\bibinfo{author}{Ho, J.} \& \bibinfo{author}{Salimans, T.}
\newblock \bibinfo{title}{{Classifier-Free Diffusion Guidance}}.
\newblock \emph{\bibinfo{journal}{arXiv}}  (\bibinfo{year}{2022}).

\bibitem{edgar2020new}
\bibinfo{author}{Edgar, H.} \emph{et~al.}
\newblock \bibinfo{title}{New mexico decedent image database}.
\newblock \bibinfo{howpublished}{Office of the Medical Investigator, University of New Mexico} (\bibinfo{year}{2020}).

\bibitem{loshchilov2018decoupled}
\bibinfo{author}{Loshchilov, I.} \& \bibinfo{author}{Hutter, F.}
\newblock \bibinfo{title}{Decoupled weight decay regularization} (\bibinfo{year}{2019}).
\newblock \urlprefix\url{https://openreview.net/forum?id=Bkg6RiCqY7}.

\bibitem{paszke2019pytorch}
\bibinfo{author}{Paszke, A.} \emph{et~al.}
\newblock \bibinfo{title}{{PyTorch: An Imperative Style, High-Performance Deep Learning Library}}.
\newblock \emph{\bibinfo{journal}{arXiv}}  (\bibinfo{year}{2019}).

\bibitem{liu2021swin}
\bibinfo{author}{Liu, Z.} \emph{et~al.}
\newblock \bibinfo{title}{Swin transformer: Hierarchical vision transformer using shifted windows} (\bibinfo{year}{2021}).

\bibitem{ridnik2021imagenet}
\bibinfo{author}{Ridnik, T.}, \bibinfo{author}{Ben-Baruch, E.}, \bibinfo{author}{Noy, A.} \& \bibinfo{author}{Zelnik-Manor, L.}
\newblock \bibinfo{title}{{ImageNet-21K Pretraining for the Masses}}.
\newblock \emph{\bibinfo{journal}{arXiv}}  (\bibinfo{year}{2021}).

\bibitem{sudhyadhom2020on}
\bibinfo{author}{Sudhyadhom, A.}
\newblock \bibinfo{title}{{On the molecular relationship between Hounsfield Unit (HU), mass density, and electron density in computed tomography (CT)}}.
\newblock \emph{\bibinfo{journal}{PLoS One}} \textbf{\bibinfo{volume}{15}}, \bibinfo{pages}{e0244861} (\bibinfo{year}{2020}).

\bibitem{wang2012measurement}
\bibinfo{author}{Wang, C.-M.} \& \bibinfo{author}{Chen, W.-Y.}
\newblock \bibinfo{title}{The human-height measurement scheme by using image processing techniques} (\bibinfo{year}{2012}).

\end{thebibliography}

\end{document}